%% file: main.tex
\documentclass{article}
\usepackage[preprint]{neurips_2026}

\usepackage[utf8]{inputenc} 
\usepackage[T1]{fontenc}    
\usepackage{url}            
\usepackage{amsfonts}       
\usepackage{nicefrac}       
\usepackage{microtype}      
\usepackage{xcolor}         
\usepackage{booktabs,multirow,array,makecell}
\usepackage{graphicx}
\usepackage{titletoc}
\input{def}

\definecolor{nipsblue}{rgb}{0.21,0.49,0.74}
\usepackage[pagebackref,breaklinks,colorlinks,allcolors=nipsblue]{hyperref}
\title{Hear, Localize, and Reason: Spatially Aware \\Scene Understanding for Audio-Visual LLMs}

\author{
    Kim Sung-Bin$^{1}$ \quad Lee Jung-Mok$^{2}$ \quad Jinwoo Jung$^{2}$ \quad Oh Hyun-Bin$^{1}$ \\ \textbf{Hyeonggon Ryu}$^{3}$ \quad \textbf{David Harwath}$^{4}$ \quad \textbf{Tae-Hyun Oh}$^{2}$ \\ \\
    $^{1}$POSTECH \quad $^{2}$KAIST \quad $^{3}$Hankuk University of Foreign Studies \\ $^{4}$The University of Texas at Austin \\
}
\usepackage{capt-of}

\begin{document}

\maketitle
\vspace{-3mm}
\begin{center}
    \includegraphics[width=\textwidth]{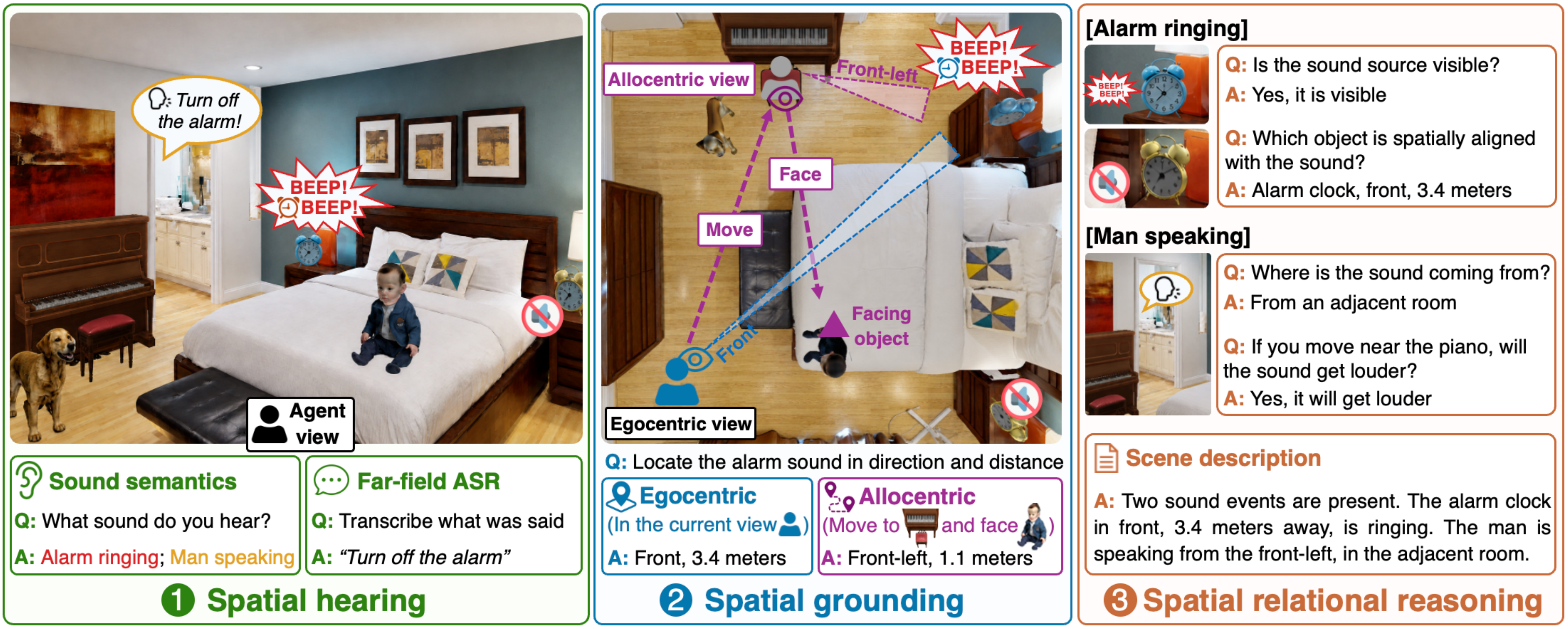}
    \vspace{-3mm}
    \captionof{figure}{
    \textbf{Overview of HLR-AVSceneQA.}
    We introduce HLR-AVSceneQA, a comprehensive benchmark for spatial audio-visual scene understanding organized around a hear-localize-reason hierarchy. For a queried sound source, the model recognizes what is heard, localizes where it originates in egocentric and allocentric views, and uses the localized source together with visual scene evidence to reason about how it relates to the surrounding scene.
    }
    \label{fig:teaser}
    \vspace{3mm}
\end{center}

\begin{abstract}
Audio-visual large language models (AV-LLMs) have made strong progress in multimodal understanding, yet most treat audio as monaural semantic content, limiting their ability to reason about where sounds originate and how they relate to the visual scene. Recent spatial audio-visual studies address parts of this problem, but often focus on specific abilities, such as spatial correspondence or direction and distance reasoning. In this work, we introduce HLR-AVSceneQA, a comprehensive benchmark with a hear-localize-reason hierarchy that evaluates whether models can recognize what is heard, localize sound sources in egocentric and allocentric views, and use these localized sources to reason about how they relate to the visual scene. We further introduce HLR-LLM, which augments a strong audio-visual foundation model with a binaural spatial audio branch alongside its monaural semantic audio pathway. We train HLR-LLM with a curriculum that incorporates final-stage chain-of-thought supervision for multi-step spatial reasoning. Experiments show that existing AV-LLMs remain limited in spatial grounding and relational reasoning, while HLR-LLM substantially improves spatial audio-visual scene understanding without compromising semantic audio-visual perception. Code and benchmark data are available at \url{https://hlrllm.github.io/}.
\end{abstract}

\input{sections/1_intro}
\input{sections/2_relatedwork}
\input{sections/3_dataset}
\input{sections/4_method}
\input{sections/5_exp}
\input{sections/6_conclusion}

\begin{ack}
This work was supported by the National Research Foundation of Korea (NRF) funded by the Korean government (MSIT) in 2026 (Grant No. RS-2024-00451947, Project Title: Interaction Framework for Connecting Users to Generative Experiences via Time-Space Expanded Multimedia: Centering on Social Communication Disorders (33\%); No. RS-2024-00358135, Corner Vision: Learning to Look Around the Corner through Multi-modal Signals (24\%)), the Institute of Information \& Communications Technology Planning \& Evaluation (IITP) grant funded by the Korean government (MSIT) (No. RS-2025-25443318, Physically-grounded Intelligence: A Dual Competency Approach to Embodied AGI through Constructing and Reasoning in the Real World (33\%); No. RS-2026-25518317, Development of AI memory mechanism that reflects human cognitive principles (10\%)), the InnoCORE program of the Ministry of Science and ICT (26-InnoCORE-01; N10250156), and the Artificial Intelligence Industrial Convergence Cluster Development Project funded by the Ministry of Science and ICT (MSIT, Korea) \& Gwangju Metropolitan City.

\paragraph{Competing interests} The authors declare no competing interests.
\end{ack}

\bibliographystyle{unsrt}
{\small
\bibliography{reference}
}
\appendix
\input{sections/7_appendix}


\end{document}

%% file: def.tex
\usepackage{enumitem}
\usepackage{multirow}
\usepackage{wrapfig}
\usepackage{colortbl}
\usepackage[normalem]{ulem}
\usepackage{microtype}
\usepackage{comment}
\usepackage[hang,flushmargin]{footmisc}
\usepackage{graphicx}
\usepackage{multicol}
\usepackage{float}
\usepackage{pifont}
\usepackage{xcolor}
\usepackage{nicematrix}
\usepackage{ulem}
\usepackage{diagbox} 
\usepackage{tipa}
\usepackage{diagbox} 
\usepackage{nicematrix}
\usepackage{makecell}
\usepackage{microtype}
\usepackage{nicematrix}   
\usepackage{booktabs}
\usepackage{xcolor}
\usepackage[table]{xcolor}
\usepackage{ulem}         
\usepackage{array}        
\usepackage{relsize}
\usepackage{tabularx}
\usepackage{booktabs}
\usepackage{multirow}
\usepackage{wrapfig}
\usepackage{amsmath,amssymb}

\setlist[itemize]{align=parleft,left=0pt,topsep=1mm,itemsep=0mm,parsep=1mm}
\definecolor{azure}{rgb}{0.0, 0.5, 1.0}
\definecolor{azure(colorwheel)}{rgb}{0.0, 0.5, 1.0}
\definecolor{nicegreen}{rgb}{0.0, 0.7, 0.1}
\definecolor{yw}{rgb}{0.01176, 0.5490, 0.5490}
\definecolor{ashblue}{rgb}{0.36, 0.54, 0.66}
\definecolor{ashgrey}{rgb}{0.7, 0.75, 0.71}
\definecolor{applegreen}{rgb}{0.55, 0.71, 0.0}
\definecolor{blue}{rgb}{0.0, 0.0, 1.0}
\definecolor{red}{rgb}{0.6, 0.0, 0.12}
\definecolor{postechred}{rgb}{0.784, 0.003, 0.313}
\definecolor{ywg}{rgb}{0.9960, 0.8984, 0.5859}
\definecolor{ballblue}{rgb}{0.13, 0.67, 0.8}
\definecolor{cornellred}{rgb}{0.7, 0.11, 0.11}
\definecolor{darkcyan}{rgb}{0.0, 0.55, 0.55}
\definecolor{CuGray}{gray}{0.9}
\definecolor{airforceblue}{rgb}{0.36, 0.54, 0.66}
\definecolor{rev}{rgb}{0.784, 0.003, 0.313}
\definecolor{pink}{cmyk}{0, 0.7808, 0.4429, 0.1412}
\definecolor{amethyst}{rgb}{0.6, 0.4, 0.8}
\definecolor{black}{rgb}{0.0, 0.0, 0.0}
\definecolor{tb3_yellow}{rgb}{0.996, 1.0, 0.6}
\definecolor{tb3_orange}{rgb}{0.980, 0.8, 0.604}
\definecolor{tb3_red}{rgb}{0.972, 0.6, 0.6}
\definecolor{dimgray}{rgb}{0.41, 0.41, 0.41}
\definecolor{brickred}{rgb}{0.8, 0.25, 0.33}
\definecolor{bleudefrance}{rgb}{0.19, 0.55, 0.91}
\definecolor{blue(ncs)}{rgb}{0.265, 0.445, 0.765}
\definecolor{blue(ryb)}{rgb}{0.01, 0.28, 1.0}
\definecolor{orange}{rgb}{1.0, 0.49, 0.0}
\definecolor{Gray}{gray}{0.88}
\definecolor{green(ncs)}{rgb}{0.0, 0.62, 0.42}
\definecolor{clova}{rgb}{0.24, 0.63, 0.33}
\definecolor{correctgreen}{rgb}{0.01, 0.702, 0.321}
\definecolor{wrongred}{rgb}{0.757, 0, 0}

\definecolor{teaserblue}{rgb}{0.274, 0.694, 1}
\definecolor{teaserorange}{rgb}{0.913, 0.443, 0.196}
\definecolor{teaserred}{rgb}{1,0,0}

\definecolor{kellygreen}{rgb}{0.3, 0.73, 0.09}

\newcolumntype{g}{>{\columncolor{CuGray}}c}
\newcolumntype{z}{>{\columncolor{CuGray}}l}

\renewcommand{\paragraph}[1]{\noindent\textbf{#1{.}}\,}

\definecolor{steelblue}{rgb}{0.27, 0.51, 0.7}

\usepackage{xspace}

\makeatletter
\def\@fnsymbol#1{\ensuremath{\ifcase#1\or *\or \dagger\or \ddagger\or
   \mathsection\or \mathparagraph\or \|\or **\or \dagger\dagger
   \or \ddagger\ddagger \else\@ctrerr\fi}}
\makeatother

\def\onedot{.\@\xspace}
 
\def\ie{\emph{i.e}\onedot}

\newcommand{\Fref}[1]{Fig.~\ref{#1}}
\newcommand{\Tref}[1]{Table~\ref{#1}}

\newcommand{\be}{\begin{eqnarray}}
\newcommand{\ee}{\end{eqnarray}}
\newcommand{\bee}{\begin{eqnarray*}}
\newcommand{\eee}{\end{eqnarray*}}

\newcommand{\matrixb}{\left[ \begin{array}}
\newcommand{\matrixe}{\end{array} \right]}

\newcommand{\cmark}{\textcolor{green!60!black}{\ding{51}}}
\newcommand{\xmark}{\textcolor{red!70!black}{\ding{55}}}
\definecolor{graygray}{HTML}{bbbbbb}
 
\usepackage{mathtools}

\definecolor{ExtFG}{HTML}{7B2CBF}   
\definecolor{OrigFG}{HTML}{A44A3F}  
\usepackage{booktabs}
\usepackage{pifont}

\usepackage{booktabs}
\usepackage{tabularx}
\usepackage{capt-of}
\usepackage{makecell}
\usepackage{ragged2e}

\newcolumntype{Y}{>{\raggedright\arraybackslash}X}
\newcolumntype{L}[1]{>{\RaggedRight\arraybackslash}p{#1}}
\newlength{\methodpanelheight}
\newcolumntype{M}[1]{>{\centering\arraybackslash}m{#1}}
\newcolumntype{C}[1]{>{\centering\arraybackslash}m{#1}}
\newcolumntype{P}[1]{>{\raggedright\arraybackslash}p{#1}}

\usepackage{tablefootnote}

%% file: sections/1_intro.tex
\section{Introduction}\label{main:intro}
Spatial audio and vision provide complementary cues for understanding a scene. 
We may hear someone calling from another room, infer that a ringing alarm clock is hidden below the bed by combining its direction with the visible layout, or decide where to move to get closer to a sound by relating auditory distance cues to nearby objects. 
Such everyday situations show that audio-visual scene understanding requires more than recognizing a sound event or matching it to a visible object. 
Accordingly, a spatially aware agent must infer where the sound originates, whether it is visible, why it may be hidden, and how its relation to the scene changes as the agent moves.

Recently, audio-visual large language models (AV-LLMs)~\cite{yao2024minicpm,cheng2024videollama,tang2025video,xu2025qwen25omnitechnicalreport,xu2025qwen3} have shown strong progress in diverse multimodal understanding tasks, such as audio-visual question answering~\cite{ye2024cat,li2022learning}, temporal grounding~\cite{chowdhury2024meerkat}, and multimodal reasoning. However, most existing AV-LLMs treat audio primarily as monaural semantic content, focusing on \textit{what} is heard rather than \textit{where} it is heard. As a result, they struggle to reason about auditory sources in the physical scene, especially when a source is outside the camera view, occluded by objects, located in another room, or spatially ambiguous among multiple candidates. Spatial audio, such as binaural or multi-channel recordings, offers the direction and distance cues needed to overcome this limitation and support spatially grounded scene understanding.

Despite this growing need for spatially grounded audio-visual understanding, two key challenges remain. First, existing spatial audio-(visual) benchmarks typically focus on 
specific subproblems,
such as audio-only~\cite{zheng2024bat,biswas2025owl,devnani2024learning} or speech-centric settings~\cite{tang2024can}, egocentric or allocentric direction and distance reasoning~\cite{chen2025savvy}, or precise RGB-D-based 3D grounding~\cite{liu2026jaeger}. 
Existing evaluations therefore remain fragmented across modalities, assumptions, and task formats, making it difficult to assess a single AV-LLM's ability to jointly handle semantic understanding, spatial grounding, and relational reasoning.
Second, although spatial audio-visual tasks require multi-step reasoning, 
many
current training settings rely on 
direct-answer
supervision~\cite{ryu2025hear,liu2026jaeger}. This gives models insufficient structure for decomposing complex spatial queries into interpretable intermediate steps that connect auditory semantics, spatial cues, and visual context.

In this work, we introduce \textbf{HLR-AVSceneQA}, a comprehensive benchmark for spatial audio-visual scene understanding. As shown in~\Fref{fig:teaser}, HLR-AVSceneQA follows a hear-localize-reason hierarchy: spatial hearing evaluates what is heard, spatial grounding evaluates source localization in egocentric and allocentric views, and spatial relational reasoning evaluates how localized sound sources relate to the visual scene.
By treating localized sound-source information not only as a prediction target but also as spatial evidence for downstream reasoning, HLR-AVSceneQA enables systematic evaluation of spatial grounding and scene-level reasoning.
Unlike panoramic-input settings~\cite{ryu2025hear}, HLR-AVSceneQA uses normal-field-of-view observations, requiring models to reason about visible, occluded, and off-screen sources from partial visual evidence. We construct the benchmark with a simulator-based pipeline~\cite{chen2022soundspaces}, producing diverse, controllable spatial audio-visual scenes with reliable annotations.

We further introduce \textbf{HLR-LLM}, a spatially aware AV-LLM that extends audio-visual semantic understanding with explicit spatial audio modeling. 
Built on an existing AV-LLM~\cite{xu2025qwen25omnitechnicalreport}
and augmented with a spatial audio branch, HLR-LLM 
improves spatial grounding and reasoning while preserving strong semantic perception.
To address the multi-step nature of spatial audio-visual reasoning, we incorporate chain-of-thought~\cite{wei2022chain,zhang2023multimodal} supervision in the final stage of curriculum training, encouraging the model to connect auditory, spatial, and visual cues rather than directly predicting an answer.

Experiments show that existing AV-LLMs, including commercial models that rely on monaural audio, struggle with spatial grounding and reasoning in our benchmark.
In contrast, HLR-LLM preserves strong semantic hearing ability, including sound event and speech recognition, while substantially improving spatial grounding and relational reasoning. 
Transfer experiments on real-world spatial audio-visual recordings~\cite{shimada2023starss23} further suggest that pretraining on HLR-AVSceneQA provides useful spatial priors for adaptation to real-world data.
Our contributions are summarized as follows:

\begin{itemize}
    \item We introduce HLR-AVSceneQA, a comprehensive benchmark that evaluates spatial hearing and grounding both directly and as intermediate evidence for downstream spatial relational reasoning.

    \item We propose HLR-LLM, a spatially aware AV-LLM that augments a strong foundation model with a spatial audio pathway to improve grounding and reasoning while preserving semantic perception.
    
    \item We demonstrate that final-stage chain-of-thought curriculum training improves multi-step spatial reasoning by connecting auditory, spatial, and visual cues through intermediate reasoning steps.
\end{itemize}

%% file: sections/2_relatedwork.tex
\section{Related work}\label{main:related}
\begin{table}[t]
\caption{
\textbf{Evaluation scope of spatial audio and audio-visual benchmarks.}
We compare whether each benchmark explicitly evaluates key spatial audio-visual capabilities.
A+V denotes audio-visual input; Multi-source denotes simultaneous sound sources; Far-field ASR and Sound event denote speech recognition from far-field audio and non-speech sound recognition; Ego./Allo. denote egocentric/allocentric reasoning; and Corr. denotes audio-visual correspondence.
Cross-room, Hidden Source, Proximity, and AV Caption denote reasoning about across-room sources, invisible or occluded sources, nearest-object relations, and open-ended audio-visual captioning, respectively.
}\vspace{-1mm}
\label{tab:benchmark_comparison}
\centering
\small
\renewcommand{\arraystretch}{0.9}
\resizebox{1\linewidth}{!}{%
\begin{tabular}{@{\,}l@{\quad\,\,}c@{\quad\,\,}c@{\quad\,\,}c@{\quad\,\,}c@{\quad\,\,}c@{\quad\,\,}c@{\quad\,\,}c@{\quad\,\,}c@{\quad\,\,}c@{\quad\,\,}c@{\quad\,\,}c@{\,}}
\toprule
\textbf{Benchmark} &
\textbf{A+V} &
\makecell{\textbf{Multi-}\\\textbf{source}} &
\makecell{\textbf{Far-field}\\\textbf{ASR}} &
\makecell{\textbf{Sound}\\\textbf{event}} &
\textbf{Ego.} &
\textbf{Allo.} &
\makecell{\textbf{AV}\\\textbf{Corr.}} &
\makecell{\textbf{Cross-}\\\textbf{room}} &
\makecell{\textbf{Hidden}\\\textbf{Source}} &
\makecell{\textbf{Proximity}} &
\makecell{\textbf{AV}\\\textbf{Caption}} \\
\midrule
BAT~\cite{zheng2024bat}
  & \xmark & \cmark & \xmark & \cmark & \cmark & \xmark & \xmark & \xmark & \xmark & \xmark & \xmark \\
SpatialWhisper~\cite{tang2024can}
  & \xmark & \cmark & \cmark & \xmark & \cmark & \xmark & \xmark & \xmark & \xmark & \xmark & \xmark \\
Hear You Are~\cite{ryu2025hear}
  & \cmark & \xmark & \xmark & \cmark & \cmark & \xmark & \cmark & \xmark & \xmark & \xmark & \xmark \\
SAVVY-Bench~\cite{chen2025savvy}
  & \cmark & \xmark & \xmark & \xmark & \cmark & \cmark & \xmark & \cmark & \xmark & \xmark & \xmark \\
SpatialSceneQA~\cite{liu2026jaeger}
  & \cmark & \cmark & \xmark & \xmark & \cmark & \xmark & \cmark & \xmark & \xmark & \xmark & \xmark \\
\midrule
\textbf{HLR-AVSceneQA}
  & \cmark & \cmark & \cmark & \cmark & \cmark & \cmark & \cmark & \cmark & \cmark & \cmark & \cmark \\
\bottomrule
\end{tabular}%
}

\end{table}

\paragraph{Spatial audio understanding}
Spatial audio provides geometric cues about sound sources beyond monaural semantic content. While monaural audio understanding mainly targets semantic tasks, such as automatic speech recognition~\cite{baevski2020wav2vec,radford2023robust}, sound event recognition~\cite{gemmeke2017audio,kong2020panns,gong2021ast}, and audio captioning~\cite{drossos2020clotho,kim2019audiocaps,kim2023prefix}, spatial audio formats, such as binaural audio and first-order ambisonics (FOA), capture direction- and distance-related cues.
Early work on spatial audio explores sound event localization and detection (SELD), where models jointly identify active sound events and estimate the direction of arrival~\cite{adavanne2018sound,shimada2021accdoa,shimada2022multi}. More recent work connects spatial audio with language models for sound localization and reasoning~\cite{zheng2024bat,biswas2025owl, hyun2026spatio}, learns spatially aware audio-text representations through contrastive objectives~\cite{devnani2024learning,seki2025spatial}, and adapts FOA inputs for far-field speech understanding~\cite{tang2024can,xie2025thinking}. These works highlight the value of spatial hearing, but remain largely audio-only or speech-centric, leaving open how spatial auditory cues should be grounded in and reasoned over a visual scene.

\paragraph{Spatial audio-visual benchmarks and tasks}
General audio-visual benchmarks~\cite{sung2024avhbench,chowdhury2025avtrustbench,gong2024av,yang2022avqa,zhou2025daily} evaluate broad multimodal abilities, but do not focus on spatial sound-source understanding. Recent benchmarks have begun to address specific spatial capabilities: SAVVY-Bench~\cite{chen2025savvy} evaluates dynamic egocentric/allocentric direction and distance reasoning, SpatialSceneQA~\cite{liu2026jaeger} focuses on RGB-D and FOA-based 3D grounding, and Hear You Are~\cite{ryu2025hear} studies sound-object correspondence with panoramic images. As summarized in \Tref{tab:benchmark_comparison}, these benchmarks cover complementary aspects of spatial audio-visual understanding, but focus on different subsets of capabilities. HLR-AVSceneQA instead provides a unified hear-localize-reason evaluation that spans spatial hearing, grounding, multi-source scenes, and scene-level relational reasoning, including whether localized sound-source information can support broader reasoning about the surrounding scene. It further uses normal-field-of-view egocentric observations, requiring models to reason about visible, off-screen, and hidden sound sources from partial visual observations.

\paragraph{Spatial audio-visual large language models (AV-LLMs)}
General AV-LLMs~\cite{yao2024minicpm,cheng2024videollama,tang2025video,xu2025qwen25omnitechnicalreport,xu2025qwen3} support video understanding, temporal localization, and audio-visual question answering, but mostly rely on monaural audio. Spatially aware AV-LLMs add spatial auditory cues to visual inputs. Hear You Are LLM~\cite{ryu2025hear} uses spatial audio and panoramic images, SAVVY~\cite{chen2025savvy} builds dynamic spatial maps with a training-free pipeline, and JAEGER~\cite{liu2026jaeger} integrates RGB-D observations with FOA audio for 3D grounding and reasoning. While promising, these systems are often tailored to specific tasks, such as direction/distance QA or precise 3D grounding, and do not fully inherit the broad semantic and reasoning abilities of general AV-LLMs. They are also commonly trained with direct-answer supervision despite the multi-step nature of spatial audio-visual reasoning~\cite{ryu2025hear,liu2026jaeger}. HLR-LLM addresses these gaps by extending a strong AV-LLM with a spatial audio branch and chain-of-thought supervision~\cite{wei2022chain,zhang2023multimodal}, improving spatial reasoning while preserving general audio-visual capabilities.

%% file: sections/3_dataset.tex
\section{HLR-AVSceneQA: Spatial audio-visual scene understanding benchmark}\label{main:dataset}

\begin{table}[t]
\centering
\footnotesize
\setlength{\tabcolsep}{3pt}
\renewcommand{\arraystretch}{0.85}
\caption{
\textbf{Representative QA examples in HLR-AVSceneQA.}
Most tasks use controlled answer formats, such as multiple choice or discretized bins; simplified prompts and answers are shown for clarity. The final row shows one CoT supervision example used for rationale-augmented training.
}
\vspace{-1mm}
\label{tab:task_summary}
\begin{tabularx}{\linewidth}{@{}L{0.15\linewidth}@{\hspace{1.5pt}}L{0.14\linewidth}Y@{}}
\toprule
\textbf{Task family} & \textbf{Subtask} & \textbf{Representative example} \\
\midrule

\multirow{2}{*}[0.5ex]{\makecell[c]{\textbf{Spatial}\\\textbf{hearing}}}
& Sound event
& \textbf{\textit{Q}:} What sound do you hear? \quad \textbf{\textit{A}:} \texttt{Hair dryer drying}. \\

& Far-field ASR
& \textbf{\textit{Q}:} Write the exact phrase spoken. \quad \textbf{\textit{A}:} \texttt{``Yes it is suppressed''}. \\

\midrule

\multirow{2}{*}[-0.8ex]{\makecell[c]{\textbf{Egocentric}\\\textbf{grounding}}}
& Direction
& \textbf{\textit{Q}:} From the current viewpoint, which direction is the sound coming from?
\newline \textbf{\textit{A}:} \texttt{front-left}. \\

& Distance
& \textbf{\textit{Q}:} From the current viewpoint, how far is the sound source? \quad \textbf{\textit{A}:} \texttt{2.5--3 m}. \\

\midrule

\multirow{2}{*}[-1.9ex]{\makecell[c]{\textbf{Allocentric}\\\textbf{grounding}}}
& Direction
& \textbf{\textit{Q}:} If you stand at the sofa and face the TV, which direction is the sound coming from? \quad \textbf{\textit{A}:} \texttt{back-left}. \\

& Distance
& \textbf{\textit{Q}:} If you stand at the sofa and face the TV, how far is the sound source?
\newline \textbf{\textit{A}:} \texttt{0--1.0 m}. \\

\midrule

\multirow{6}{*}[-3.7ex]{\makecell[l]{\textbf{Spatial}\\\textbf{relational}\\\textbf{reasoning}}}
& Visibility
& \textbf{\textit{Q}:} From the current viewpoint, is the sound source visible? \quad \textbf{\textit{A}:} \texttt{no}. \\

& Hidden-source
& \textbf{\textit{Q}:} Given the sound, where is its source located?
\newline \textbf{\textit{A}:} \texttt{hidden behind the armchair}. \\

& Correspondence
& \textbf{\textit{Q}:} Which object is spatially aligned with the sound? \quad \textbf{\textit{A}:} \texttt{man in front-right}. \\

& Loudness change
& \textbf{\textit{Q}:} If you move to the sofa, does the sound get louder or quieter? \quad \textbf{\textit{A}:} \texttt{quieter}. \\

& Proximity
& \textbf{\textit{Q}:} Which object among the options is closest to the sound source? \quad \textbf{\textit{A}:} \texttt{armchair}. \\

& Scene captioning
& \textbf{\textit{Q}:} Describe the sound event(s), including direction, visibility, and hidden location if applicable.
\newline \textbf{\textit{A}:} \texttt{A hair dryer is drying from the front-left direction. It is not visible and is likely occluded by the armchair.} \\

\midrule

\multicolumn{2}{@{}L{0.29\linewidth}}{%
\begin{minipage}[t]{\linewidth}
\raggedright
\textbf{CoT supervision example}\\[-0.15em]
\textit{provided for allocentric grounding and spatial relational reasoning;}\\[-0.15em]
\textit{hidden-source example shown}
\end{minipage}
}
&
\begin{minipage}[t]{\linewidth}
\raggedright
\textbf{\textit{Q}:} Given the sound, where is its source located?

\vspace{0.15em}
\textbf{\textit{A}:} {\ttfamily The sound event is a hair dryer drying, arriving from the front-left direction and 2.7 meters away. The hair dryer is not visible because it is occluded by the armchair. Therefore, the source is hidden behind the armchair.}
\end{minipage}
\\

\bottomrule
\end{tabularx}
\end{table}

We introduce HLR-AVSceneQA, a comprehensive benchmark for spatial audio-visual scene understanding. HLR-AVSceneQA follows a hear-localize-reason hierarchy that evaluates three capabilities: \textit{spatial hearing} for recognizing what is heard, \textit{spatial grounding} for localizing sound sources from egocentric and allocentric viewpoints, and \textit{spatial relational reasoning} for understanding how localized sources relate to visible, hidden, nearby, or off-screen objects in the scene (\Fref{fig:teaser}). This hierarchy spans 12 tasks, as summarized in \Tref{tab:task_summary}. HLR-AVSceneQA contains 175,061 training and 21,808 test QA pairs, including 53,279 training and 4,869 test pairs from scenes with multiple simultaneously active sound sources. We also provide 94,519 CoT-augmented training pairs for multi-step spatial reasoning. Because collecting real-world spatial audio-visual data with accurate annotations is challenging, we construct the benchmark with a two-stage simulator-based pipeline: Stage I synthesizes spatial audio-visual scenes with structured metadata, and Stage II generates QA pairs from this metadata.

\subsection{Stage I: Spatial audio-visual scene synthesis}
\begin{figure}[t]
    \centering
    \includegraphics[width=1\linewidth]{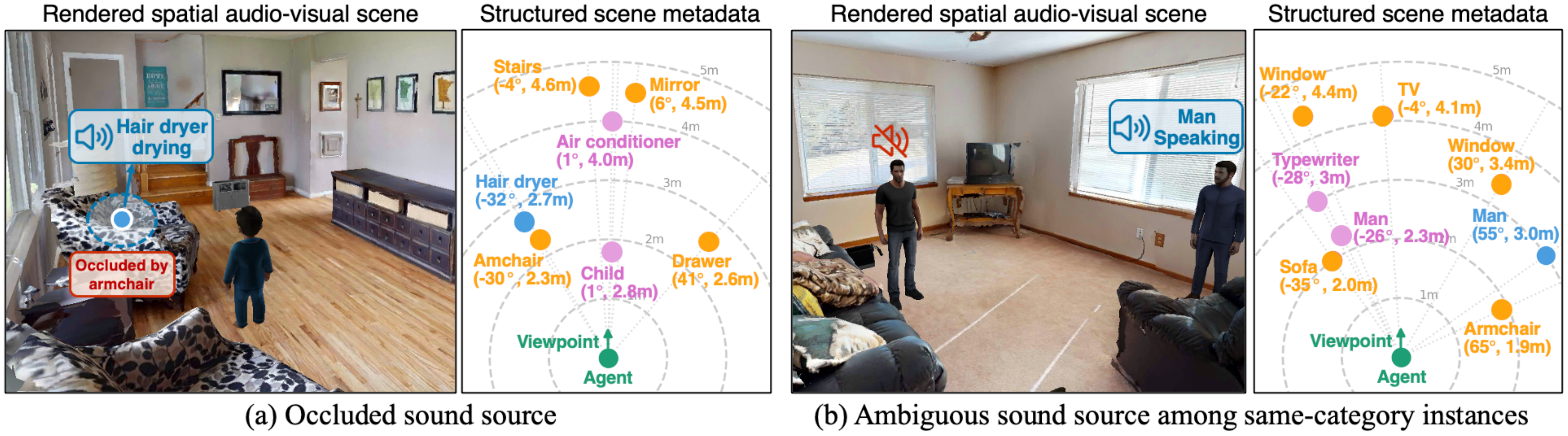}
    \vspace{-6mm}
    \caption{\textbf{Examples of rendered scenes and metadata.} Our dataset covers diverse scenarios that require spatially aware audio-visual reasoning, including (a) sound sources occluded by objects and (b) ambiguous sound sources among same-category instances. Each sample is summarized with structured scene metadata. \textcolor[HTML]{ffa50a}{Yellow} denotes existing scene objects, \textcolor[HTML]{e59edd}{pink} denotes inserted objects, \textcolor[HTML]{499ee0}{blue} denotes sounding objects, and \textcolor[HTML]{1c9e75}{green} denotes the agent.}
    \label{fig:dataexamples}
    \vspace{-1mm}
\end{figure}

We synthesize each sample in an indoor 3D scene from an agent's egocentric viewpoint, where the agent pose defines both the RGB camera and the acoustic receiver. 
We populate the scene with diverse candidate objects and designate up to two objects as active sound sources. Each sample includes a synchronized RGB observation, binaural audio, object layouts, source assignments, and geometric metadata for QA generation. Examples of rendered scenes and metadata are shown in \Fref{fig:dataexamples}.

\paragraph{Scene and acoustic simulator}
We use SoundSpaces 2.0~\cite{chen2022soundspaces}, which renders geometry-aware room acoustics and spatializes dry monaural clips into binaural audio. For 3D environments, we use HM3DSem~\cite{yadav2023habitat}, which provides 216 indoor spaces and 3,100 rooms with dense object- and room-level semantics. These semantics support realistic rendering, layout-aware QA generation, and filtering of physically implausible placements. We use 145 scenes for training and 36 for testing.

\paragraph{Audio source curation}
We use non-speech sounds from VGGSound~\cite{chen2020vggsound}, which contains over 200,000 in-the-wild clips across 310+ classes. 
Since some classes, such as \textit{sea waves} or \textit{people cheering}, are not naturally grounded to a single indoor object, we manually exclude them.
We further use ImageBind-based text-audio similarity to filter out likely mislabeled clips~\cite{girdhar2023imagebind}. For speech audio, we use LibriSpeech~\cite{panayotov2015librispeech}, which provides clean utterance-level transcripts for far-field speech recognition. We follow the official train/test splits of both VGGSound and LibriSpeech.

\paragraph{Visual object curation}
We build a 3D mesh pool of sounding objects to insert into scenes. We collect available meshes from Objaverse-XL~\cite{deitke2023objaverse}, and supplement categories with insufficient coverage using off-the-shelf 2D/3D generation models~\cite{rombach2022high,xiang2025structured}. 
All assets are manually checked to remove low-quality meshes. We keep approximately 81 meshes per category for training and 35 for testing.

\paragraph{Spatial audio-visual scene assembly and structured metadata}
Given a room from HM3DSem, we insert 2 to 6 candidate object meshes based on available free space and room scale. We activate up to two sound sources and attach them to selected objects. Some audio-object pairs are semantically matched, while others are intentionally \textit{spatially aligned but semantically mismatched}. We further include sources that are fully visible, occluded by an object, placed in an adjacent room, or outside the agent's field of view, introducing realistic ambiguity and preventing reliance on category matching alone. The final RGB image is rendered from the agent viewpoint with a 90$^\circ$ horizontal field of view at 1280$\times$720 resolution, paired with binaural audio. For each sample, we export structured metadata, including agent pose, object identities and 3D coordinates, source-object assignments, visibility state, occlusion type, and pairwise geometry, to generate QA pairs with accurate labels.

\subsection{Stage II: Benchmark task design and QA construction}
Given each rendered scene and its metadata, we generate QA pairs for three task families. To improve linguistic diversity while keeping evaluation stable, we vary question templates but preserve controlled answer formats. Most tasks use constrained outputs, such as multiple choice or discretized bins, while speech transcription and scene captioning are open-ended. Representative examples are shown in Table~\ref{tab:task_summary}, and additional construction and quality-control details are provided in Appendix~\ref{app:detail}.

\paragraph{Task family 1: Spatial hearing}
This family evaluates semantic perception from spatial audio through \textbf{\textit{sound event recognition}} and \textbf{\textit{far-field speech recognition}}, which identify audible event categories and transcribe spoken content as the foundation for grounding and reasoning.

\paragraph{Task family 2: Spatial grounding}
This family evaluates sound-source localization in egocentric and allocentric frames, each with \textbf{\textit{direction}} and \textbf{\textit{distance}} estimation. Egocentric grounding localizes the source relative to the agent's current viewpoint, while allocentric grounding localizes it from a transformed viewpoint, defined by standing at a reference object and facing another object. This tests both source localization and viewpoint-aware spatial reasoning.

\paragraph{Task family 3: Spatial relational reasoning}
This family extends source localization to scene-level reasoning by evaluating how localized sound sources relate to the visual scene. \textbf{\textit{Visibility}} asks whether the source is 
visible from the current view,
and \textbf{\textit{correspondence}} asks which visible object is spatially aligned with the sound, where the answer may be semantically mismatched or \texttt{None}. \textbf{\textit{Hidden-source localization}} identifies whether a non-visible source is hidden behind an object, in an adjacent room, or outside the field of view. \textbf{\textit{Loudness change}} asks whether the source becomes louder or quieter after the agent moves to a target object; \textbf{\textit{proximity reasoning}} asks which candidate object is nearest to the source; and \textbf{\textit{scene captioning}} summarizes sound events, directions, and visibility states, including multiple emitters when present.

\paragraph{Chain-of-thought supervision}
We further provide chain-of-thought (CoT) answer variants for allocentric grounding and spatial relational reasoning. 
The CoT rationales are deterministically generated from structured simulator metadata using task-specific templates, while GPT-4~\cite{openai2023gpt4} is used only to paraphrase question wording.
These rationales expose task-specific intermediate cues beyond the final answer, such as sound event identity, source location, visibility checks, distance comparisons, and how source direction and distance change under transformed viewpoints. 
This supervision encourages the model to connect auditory, spatial, and visual cues through explicit intermediate reasoning steps rather than directly predicting the final answer.

%% file: sections/4_method.tex
\begin{figure}[t]
    \centering
    \setlength{\methodpanelheight}{0.185\textheight}

    \begin{minipage}[t]{0.56\linewidth}
        \vspace{0pt}
        \centering
        \parbox[c][\methodpanelheight][c]{\linewidth}{
            \centering
            \includegraphics[
                width=\linewidth,
                height=\methodpanelheight,
                keepaspectratio
            ]{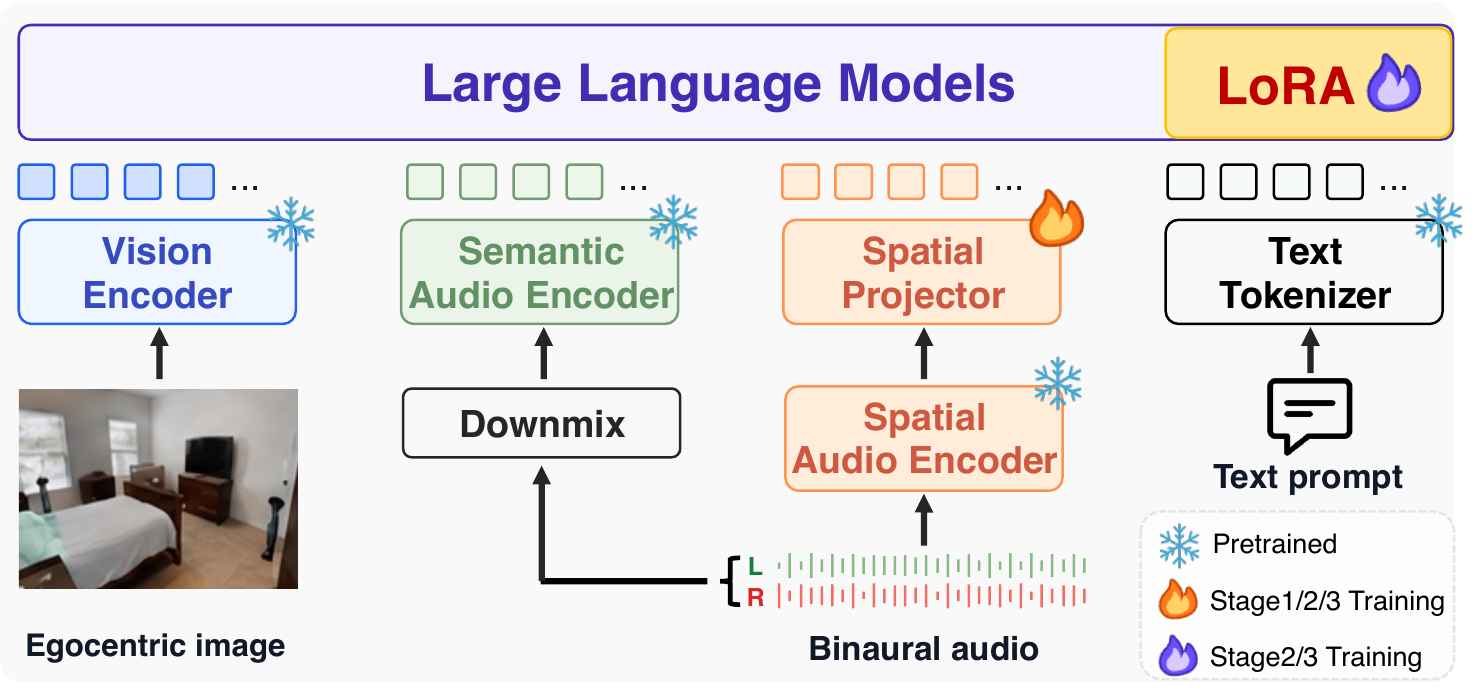}
        }
    \end{minipage}
    \hfill
    \begin{minipage}[t]{0.40\linewidth}
        \vspace{0pt}
        \centering
        \parbox[c][\methodpanelheight][c]{\linewidth}{
            \centering
            \fontsize{6.4}{7.4}\selectfont
            \setlength{\tabcolsep}{2.2pt}
            \renewcommand{\arraystretch}{1.18}

            \begin{tabularx}{\linewidth}{@{}cYc@{}}
                \toprule
                \textbf{Stage} &
                \textbf{Training Dataset $\rightarrow$\,Objective} &
                \textbf{\# Samples} \\
                \midrule

                \multirow[c]{2}{*}{1} &
                Spatial audio-only QA~\cite{zheng2024bat} &
                \multirow[c]{2}{*}{514k} \\
                &
                $\rightarrow$\,Spatial audio alignment &
                \\
                \midrule

                \multirow[c]{2}{*}{2} &
                HLR-AVSceneQA &
                \multirow[c]{2}{*}{175k} \\
                &
                $\rightarrow$\,Spatial audio-visual instruction tuning &
                \\
                \midrule

                \multirow[c]{2}{*}{3} &
                HLR-AVSceneQA-CoT &
                \multirow[c]{2}{*}{94.5k} \\
                &
                $\rightarrow$\,Chain-of-thought tuning &
                \\
                \bottomrule
            \end{tabularx}
        }
    \end{minipage}

    \vspace{-2mm}
    \makebox[0.56\linewidth][c]{\small(a) Model architecture}
    \hfill
    \makebox[0.40\linewidth][c]{\small(b) Training curriculum}
    \vspace{-1mm}
    \caption{\textbf{Architecture and training curriculum of HLR-LLM.}
    (a) HLR-LLM augments a strong AV-LLM with a binaural spatial audio branch, enabling the language model to jointly attend to text, visual, downmixed monaural audio, and binaural spatial audio tokens.
    (b) The model is trained through a three-stage curriculum: spatial audio alignment, spatial audio-visual instruction tuning, and final-stage CoT tuning for structured spatial reasoning.}
    \label{fig:method}
\end{figure}

\section{Method}\label{main:method}
We introduce HLR-LLM for spatial audio-visual scene understanding.
Our goal is to enable spatial audio-visual grounding and reasoning with binaural spatial cues while retaining the semantic perception ability of a strong base AV-LLM.
To this end, HLR-LLM introduces a spatial audio pathway to the base AV-LLM, where a spatial audio encoder and projector provide binaural spatial cues alongside visual context and monaural semantic audio.
We train the model with a curriculum that incorporates chain-of-thought supervision to guide multi-step spatial reasoning.

\subsection{HLR-LLM architecture}
Figure~\ref{fig:method}~(a) shows the architecture of HLR-LLM.
The model consists of the base visual and monaural audio encoders, a newly introduced spatial audio encoder, a lightweight spatial audio projector, and the language model.
The monaural audio branch preserves semantic hearing abilities, such as sound event recognition and speech understanding, while the spatial audio branch provides direction- and distance-sensitive cues for grounding and reasoning. 
The language model integrates visual, monaural audio, spatial audio, and text tokens to generate task-specific responses.

\paragraph{Spatial audio tokenization}
Given a binaural audio signal $b \in \mathbb{R}^{2 \times T}$, we encode it with a pretrained spatial audio encoder $f_{\mathrm{spa}}$~\cite{zheng2024bat} and map the resulting features into the language model embedding space using a trainable MLP projector $p_{\mathrm{spa}}$:
$Z_{\mathrm{spa}} = p_{\mathrm{spa}}(f_{\mathrm{spa}}(b)).$
The projected embeddings $Z_{\mathrm{spa}}$ serve as spatial audio tokens that provide acoustic localization cues to the LLM.

\paragraph{Integration with the AV-LLM}
We build HLR-LLM on Qwen2.5-Omni~\cite{xu2025qwen25omnitechnicalreport}, which supports text, image, and audio inputs. We introduce a new media token, \texttt{<spatial\_audio>}, whose placeholder positions are replaced with the projected spatial audio tokens $Z_{\mathrm{spa}}$. 
The resulting multimodal input sequence is
$X = [E(q), Z_{\mathrm{vis}}, Z_{\mathrm{aud}}, Z_{\mathrm{spa}}],$
where $E(q)$ denotes text embeddings, $Z_{\mathrm{vis}}$ visual tokens, $Z_{\mathrm{aud}}$ monaural audio tokens, and $Z_{\mathrm{spa}}$ spatial audio tokens. The language model generates the answer $y$ conditioned on this sequence, \ie, $p(y \mid X)$. 
This design preserves the base model's semantic audio-visual ability while explicitly injecting spatial acoustic cues.

\subsection{Curriculum training for spatial audio-visual reasoning}
Spatial audio-visual scene understanding requires low-level perception of what is heard and where it comes from, as well as high-level spatial reasoning over the scene. 
Figure~\ref{fig:method}~(b) summarizes the three-stage curriculum used to train HLR-LLM, which progresses from spatial audio alignment to spatial audio-visual instruction tuning and final-stage CoT tuning.

\paragraph{Stage 1: Spatial audio alignment}
We first pretrain the spatial audio projector using a spatial audio-only QA dataset~\cite{zheng2024bat}. 
This aligns frozen spatial audio encoder features with the LLM embedding space and initializes the model to verbalize spatial audio properties, such as sound event and location.

\paragraph{Stage 2: Spatial audio-visual instruction tuning}
We then fine-tune the model on the training split of HLR-AVSceneQA. This stage teaches the model to jointly use visual tokens, monaural audio tokens, and spatial audio tokens across spatial hearing, grounding, and relational reasoning tasks. For example, recognizing a ringing sound can rely on the monaural audio branch, while determining whether the source is visible or hidden requires combining spatial cues with visual evidence.

\paragraph{Stage 3: Chain-of-thought tuning}
Finally, we continue instruction tuning with CoT-formatted answers for multi-step spatial tasks. 
The model learns to produce a compact rationale that connects auditory cues, spatial estimates, and visual evidence before giving a normalized final answer. 
This encourages 
the model
to decompose spatial audio-visual queries into interpretable reasoning steps rather than relying on direct answer prediction.

\paragraph{Training objective}
At each stage, we minimize the standard auto-regressive cross-entropy loss over the target answer tokens.
Stage 1 updates only the spatial audio projector, while Stages 2--3 update the spatial audio projector and LoRA parameters of the language model.
All visual, monaural audio, and spatial audio encoders remain frozen throughout training.

\subsection{Implementation details}
We follow the audio preprocessing of BAT~\cite{zheng2024bat}. Each source clip is loudness-normalized and convolved with the room impulse response generated by SoundSpaces to produce binaural audio. The resulting two-channel waveform is sampled at 32\,kHz. We use a window size of 1024, a hop size of 320, and 128 mel bins to compute spectrogram features. The spatial audio branch receives the binaural signal, while the monaural audio branch receives its channel-averaged version. For visual and monaural audio inputs, we use the default preprocessing of Qwen2.5-Omni without modification.
For LoRA fine-tuning, we use rank 16, alpha 16, dropout 0.05, a learning rate of $2\times10^{-4}$, and weight decay of 0.01. We train for 2 epochs on four NVIDIA A6000 GPUs with a per-device batch size of 4.

%% file: sections/5_exp.tex
\section{Experiment}\label{main:exp}
\subsection{Experimental setup}
\paragraph{Datasets and metrics}
We primarily evaluate models on HLR-AVSceneQA for comprehensive spatial audio-visual scene understanding. Except for far-field ASR, which is evaluated by word error rate (WER), and scene captioning, which is evaluated with CIDEr-D~\cite{vedantam2015cider}, METEOR~\cite{banerjee2005meteor}, and SPIDEr~\cite{liu2017improved}, all other tasks are evaluated using accuracy. We also construct QA formats on the STARSS23~\cite{shimada2023starss23} dataset to evaluate adaptation to real-world spatial audio-visual recordings, covering sound event recognition, egocentric direction and distance prediction, and visibility prediction. 
We restrict the evaluation to these tasks because STARSS23 does not provide the object- and scene-level annotations required to reliably define hidden-source, allocentric, and other relational reasoning tasks.
All fine-tuned models are trained on the corresponding training split and evaluated on the test split. Unless otherwise noted, we report mean $\pm$ standard deviation over three runs.

\paragraph{Competing methods}
We compare HLR-LLM with both zero-shot and fine-tuned audio-visual models. For zero-shot evaluation, we consider several open-source AV-LLMs~\cite{yao2024minicpm,cheng2024videollama,tang2025video,xu2025qwen25omnitechnicalreport,xu2025qwen3} and commercial models~\cite{openai_realtime_api_docs,comanici2025gemini}, which mainly use visual input and monaural audio. We also compare with SAVVY~\cite{chen2025savvy}, which uses multichannel audio but focuses on spatial grounding tasks. For fine-tuned baselines, we initialize Hear You Are LLM~\cite{ryu2025hear} and Qwen2.5-Omni~\cite{xu2025qwen25omnitechnicalreport} from their respective public checkpoints and fine-tune them on the HLR-AVSceneQA training split. Although Hear You Are LLM is originally trained with panoramic images, we adapt it to the same normal-FOV inputs and HLR-AVSceneQA supervision by fine-tuning its visual pathway, projectors/adapters, and LLM while keeping the audio encoder frozen. We additionally fine-tune a dual-channel variant of Qwen2.5-Omni, which separately encodes the left and right binaural channels through the original audio pathway and feeds both audio token streams to the LLM without an explicit spatial audio branch.

\subsection{Main results on HLR-AVSceneQA}
Tables~\ref{tab:model_sem_ground} and~\ref{tab:model_full} summarize the main results on HLR-AVSceneQA.
In \Tref{tab:model_sem_ground}, zero-shot AV-LLMs and commercial models show reasonable spatial hearing ability, but perform poorly on direction and distance grounding, indicating that monaural audio with visual input is insufficient for spatial localization.
Baselines with spatial audio improve some grounding tasks, with SAVVY~\cite{chen2025savvy} performing well on egocentric direction and Hear You Are LLM improving over zero-shot models, but they remain limited in broader grounding and reasoning.
Fine-tuning Qwen2.5-Omni, our base model, substantially improves sound-event recognition and spatial grounding, yet accurate localization remains challenging without explicit binaural cues.
Dual-channel Qwen2.5-Omni further improves localization by processing left and right channels separately through the original audio pathway, confirming the value of binaural spatial cues.
HLR-LLM outperforms this baseline across all egocentric and allocentric grounding metrics, demonstrating the benefit of explicitly modeling binaural spatial information rather than exposing the LLM to separate left- and right-channel tokens.

\begin{table}[t]
\centering
\small
\renewcommand{\arraystretch}{0.80}
\caption{\textbf{Evaluation on spatial hearing and spatial grounding tasks.}
Sound event recognition and grounding tasks are reported in accuracy (\%), while speech recognition is reported in WER (\%, lower is better). 
\textbf{V}, \textbf{M}, and \textbf{B} denote visual input, monaural audio, and binaural audio, respectively.
\textbf{M}\textsubscript{L} and \textbf{M}\textsubscript{R} denote independently encoded left- and right-channel audio inputs.
\textbf{Best} and \uline{second-best} mean results are shown in bold and underlined, respectively.}
\vspace{-1.5mm}
\resizebox{\linewidth}{!}{%
\begin{tabular}{@{}l c cc cc cc@{}}
\toprule
\multirow{2}{*}{\textbf{Model}} 
& \multirow{2}{*}{\textbf{Modality}}
& \multicolumn{2}{c}{\textbf{Spatial Hearing}}
& \multicolumn{2}{c}{\textbf{Egocentric Grounding}}
& \multicolumn{2}{c}{\textbf{Allocentric Grounding}} \\
\cmidrule(lr){3-4}\cmidrule(lr){5-6}\cmidrule(lr){7-8}
& 
& \makecell{Sound (Acc. \%)}
& \makecell{Speech (WER $\downarrow$)}
& Direction
& Distance
& Direction
& Distance \\
\midrule
\textcolor{gray}{Random chance}
& 
& \textcolor{gray}{10.00} 
& \textcolor{gray}{-} 
& \textcolor{gray}{12.50} 
& \textcolor{gray}{10.00} 
& \textcolor{gray}{12.50} 
& \textcolor{gray}{10.00} \\
\midrule

\multicolumn{8}{l}{\textbf{\textit{Zero-shot models}}} \\
\midrule
MiniCPM-o-v2.6~\cite{yao2024minicpm}
& \textbf{V}+\textbf{M} 
& 35.26{\tiny$\pm$0.32} & 126.10{\tiny$\pm$5.51} 
& 12.83{\tiny$\pm$0.48} & 14.26{\tiny$\pm$0.66}
& 11.92{\tiny$\pm$0.36} & 18.42{\tiny$\pm$0.51} \\

VideoLLaMA2~\cite{cheng2024videollama}
& \textbf{V}+\textbf{M}
& 54.54{\tiny$\pm$0.11} & 113.54{\tiny$\pm$0.82} 
& 12.15{\tiny$\pm$0.41} & 7.73{\tiny$\pm$0.05}
& 15.84{\tiny$\pm$0.36} & 7.19{\tiny$\pm$0.17} \\

Video-SALMONN 2~\cite{tang2025video}
& \textbf{V}+\textbf{M}
& 32.20{\tiny$\pm$0.13} & $\mathbf{13.06}${\tiny$\pm$0.02}
& 17.00{\tiny$\pm$0.22} & 7.62{\tiny$\pm$0.07} 
& 17.55{\tiny$\pm$0.20} & 6.98{\tiny$\pm$0.12} \\

Qwen2.5-Omni~\cite{xu2025qwen25omnitechnicalreport}
& \textbf{V}+\textbf{M} 
& 57.21{\tiny$\pm$0.09} & 14.92{\tiny$\pm$0.56}
& 17.33{\tiny$\pm$0.07} & 7.10{\tiny$\pm$0.07}
& 21.43{\tiny$\pm$0.60} & 7.03{\tiny$\pm$0.13} \\

Qwen3-Omni~\cite{xu2025qwen3}
& \textbf{V}+\textbf{M} 
& 55.84{\tiny$\pm$0.03} & 15.27{\tiny$\pm$1.29}
& 15.64{\tiny$\pm$0.07} & 11.20{\tiny$\pm$0.06}
& 18.09{\tiny$\pm$0.21} & 9.17{\tiny$\pm$0.26} \\

SAVVY~\cite{chen2025savvy} 
& \textbf{V}+\textbf{M}+\textbf{B} 
& - & - 
& 52.36{\tiny$\pm$0.46} & 19.87{\tiny$\pm$0.15} 
& 17.81{\tiny$\pm$0.38} & 12.37{\tiny$\pm$0.60} \\

GPT-Realtime-Mini~\cite{openai_realtime_api_docs}
& \textbf{V}+\textbf{M} 
& 34.21{\tiny$\pm$0.36} & 18.07{\tiny$\pm$0.86}
& 12.65{\tiny$\pm$0.42} & 14.55{\tiny$\pm$0.60} 
& 14.27{\tiny$\pm$0.28} & 17.28{\tiny$\pm$0.96} \\

Gemini-2.5-Flash-Lite~\cite{comanici2025gemini}
& \textbf{V}+\textbf{M} 
& 38.08{\tiny$\pm$0.10} & 21.58{\tiny$\pm$0.18} 
& 19.67{\tiny$\pm$0.14} & 12.23{\tiny$\pm$0.24} 
& 19.39{\tiny$\pm$0.29} & 8.04{\tiny$\pm$0.11} \\

\midrule
\multicolumn{8}{l}{\textbf{\textit{Fine-tuned models}}} \\
\midrule
Hear You Are LLM~\cite{ryu2025hear}
& \textbf{V}+\textbf{B} 
& 67.67{\tiny$\pm$0.37} & 148.00{\tiny$\pm$1.13} 
& 44.65{\tiny$\pm$0.42} & 22.63{\tiny$\pm$0.54} 
& 21.53{\tiny$\pm$0.19} & 21.57{\tiny$\pm$0.33} \\

Qwen2.5-Omni~\cite{xu2025qwen25omnitechnicalreport}
& \textbf{V}+\textbf{M} 
& 80.57{\tiny$\pm$0.27}
& 15.33{\tiny$\pm$2.33}
& 32.69{\tiny$\pm$0.26}
& 30.92{\tiny$\pm$0.42}
& 26.26{\tiny$\pm$0.18}
& 27.73{\tiny$\pm$1.22} \\

Dual-channel Qwen2.5-Omni
& \textbf{V}+\textbf{M}\textsubscript{L}+\textbf{M}\textsubscript{R}
& $\mathbf{84.52}${\tiny$\pm$0.07}
& $\underline{13.42}${\tiny$\pm$1.14}
& $\underline{64.27}${\tiny$\pm$0.32}
& $\underline{46.59}${\tiny$\pm$0.27}
& $\underline{42.71}${\tiny$\pm$0.25}
& $\underline{37.84}${\tiny$\pm$0.24} \\

HLR-LLM (Ours)
& \textbf{V}+\textbf{M}+\textbf{B} 
& $\underline{81.03}${\tiny$\pm$0.11}
& 14.43{\tiny$\pm$0.54}
& $\mathbf{78.21}${\tiny$\pm$0.34}
& $\mathbf{55.56}${\tiny$\pm$0.27}
& $\mathbf{52.03}${\tiny$\pm$0.26}
& $\mathbf{45.08}${\tiny$\pm$0.85} \\

\bottomrule
\end{tabular}%
}
\label{tab:model_sem_ground}
\end{table}

\begin{table}[t]
\centering
\small
\caption{\textbf{Evaluation on spatial relational reasoning and scene captioning tasks.}
Relational reasoning tasks are reported in accuracy (\%). Scene captioning is evaluated with CIDEr-D, METEOR, and SPIDEr.
\textbf{V}, \textbf{M}, and \textbf{B} denote visual input, monaural audio, and binaural audio, respectively.
\textbf{M}\textsubscript{L} and \textbf{M}\textsubscript{R} denote independently encoded left- and right-channel audio inputs.
\textbf{Best} and \uline{second-best} mean results are shown in bold and underlined, respectively.}
\vspace{-1.5mm}
\resizebox{1\linewidth}{!}{%
\begin{tabular}{@{}l@{\,\,\,}c@{\,\,\,}c@{\,\,\,}c@{\,\,\,}c@{\,\,\,}c@{\,\,\,}c c@{\,\,\,}c@{\,\,\,}c@{}}
\toprule
\multirow{2}{*}{\textbf{Model}}
  & \multirow{2}{*}{\textbf{Modality}}
  & \multicolumn{5}{c}{\textbf{Spatial Relational Reasoning (Acc. \%)}}
  & \multicolumn{3}{c}{\textbf{Scene Captioning}} \\
\cmidrule(lr){3-7} \cmidrule(lr){8-10}
  & 
  & \makecell{Visibility}
  & \makecell{Correspondence}
  & \makecell{Hidden source}
  & \makecell{Loudness}
  & \makecell{Proximity}
  & CIDEr-D
  & METEOR
  & SPIDEr \\
\midrule
\textcolor{gray}{Random chance}
  & 
  & \textcolor{gray}{50.00} & \textcolor{gray}{9.49}
  & \textcolor{gray}{20.00} & \textcolor{gray}{50.00} & \textcolor{gray}{28.68}
  & \textcolor{gray}{-} & \textcolor{gray}{-}
  & \textcolor{gray}{-} \\
\midrule

\multicolumn{10}{l}{\textbf{\textit{Zero-shot models}}} \\
\midrule

MiniCPM-o-v2.6~\cite{yao2024minicpm} & \textbf{V}+\textbf{M}
& 75.20{\tiny$\pm$0.14} & 22.52{\tiny$\pm$1.09} & 11.65{\tiny$\pm$0.62} & 47.29{\tiny$\pm$0.48} & 26.83{\tiny$\pm$0.62}
& 0.078{\tiny$\pm$0.004} & 0.173{\tiny$\pm$0.001} & 0.136{\tiny$\pm$0.002} \\

VideoLLaMA2~\cite{cheng2024videollama} & \textbf{V}+\textbf{M}
& 68.37{\tiny$\pm$0.22} & 19.77{\tiny$\pm$0.14} & 21.25{\tiny$\pm$0.26} & 39.18{\tiny$\pm$0.20} & 25.52{\tiny$\pm$0.30}
& 0.112{\tiny$\pm$0.002} & 0.131{\tiny$\pm$0.001} & 0.079{\tiny$\pm$0.000} \\

Video-SALMONN 2~\cite{tang2025video} & \textbf{V}+\textbf{M}
& 75.56{\tiny$\pm$0.00} & 7.97{\tiny$\pm$0.22} & 22.00{\tiny$\pm$0.31} & 51.89{\tiny$\pm$0.00} & 30.10{\tiny$\pm$0.05}
& 0.075{\tiny$\pm$0.001} & 0.094{\tiny$\pm$0.001} & 0.074{\tiny$\pm$0.000} \\

Qwen2.5-Omni~\cite{xu2025qwen25omnitechnicalreport} & \textbf{V}+\textbf{M}
& 75.51{\tiny$\pm$0.05}
& 33.30{\tiny$\pm$0.13}
& 21.48{\tiny$\pm$0.54}
& 51.87{\tiny$\pm$0.04}
& 32.95{\tiny$\pm$0.10}
& 0.301{\tiny$\pm$0.002}
& 0.168{\tiny$\pm$0.001}
& 0.172{\tiny$\pm$0.001} \\

Qwen3-Omni~\cite{xu2025qwen3} & \textbf{V}+\textbf{M}
& 73.90{\tiny$\pm$0.30}
& 26.18{\tiny$\pm$0.28}
& 23.58{\tiny$\pm$0.36}
& 51.29{\tiny$\pm$0.85}
& 32.06{\tiny$\pm$0.39}
& 0.344{\tiny$\pm$0.004}
& 0.203{\tiny$\pm$0.000}
& 0.260{\tiny$\pm$0.001} \\

GPT-Realtime-Mini~\cite{openai_realtime_api_docs} & \textbf{V}+\textbf{M}
& 69.46{\tiny$\pm$0.61} & 12.84{\tiny$\pm$0.25} & 20.10{\tiny$\pm$0.42} & 50.13{\tiny$\pm$0.99} & 26.03{\tiny$\pm$0.44}
& 0.052{\tiny$\pm$0.002} & 0.237{\tiny$\pm$0.001} & 0.151{\tiny$\pm$0.002} \\

Gemini-2.5-Flash-Lite~\cite{comanici2025gemini} & \textbf{V}+\textbf{M}
& 49.81{\tiny$\pm$0.32} & 20.29{\tiny$\pm$0.05} & 15.36{\tiny$\pm$0.17} & 48.27{\tiny$\pm$0.57} & 30.89{\tiny$\pm$0.34}
& 0.570{\tiny$\pm$0.010} & 0.209{\tiny$\pm$0.002} & 0.218{\tiny$\pm$0.003} \\

\midrule
\multicolumn{10}{l}{\textbf{\textit{Fine-tuned models}}}\\
\midrule

Hear You Are LLM~\cite{ryu2025hear} & \textbf{V}+\textbf{B}
& 71.31{\tiny$\pm$0.05} & 54.73{\tiny$\pm$0.08} & 43.97{\tiny$\pm$0.20} & 65.41{\tiny$\pm$0.28}
& 33.33{\tiny$\pm$0.49}
& 4.160{\tiny$\pm$0.008} & 0.385{\tiny$\pm$0.000} & 2.409{\tiny$\pm$0.004} \\

Qwen2.5-Omni~\cite{xu2025qwen25omnitechnicalreport} & \textbf{V}+\textbf{M}
& 62.72{\tiny$\pm$0.10}
& 55.73{\tiny$\pm$0.07}
& 51.24{\tiny$\pm$0.58}
& 68.84{\tiny$\pm$0.32}
& 47.11{\tiny$\pm$0.27}
& \underline{4.796}{\tiny$\pm$0.014}
& 0.420{\tiny$\pm$0.000}
& 2.751{\tiny$\pm$0.007} \\

Dual-channel Qwen2.5-Omni
& \textbf{V}+\textbf{M}\textsubscript{L}+\textbf{M}\textsubscript{R}
& $\underline{77.30}${\tiny$\pm$0.12}
& $\underline{69.07}${\tiny$\pm$0.15}
& $\underline{60.12}${\tiny$\pm$0.43}
& $\underline{72.59}${\tiny$\pm$0.31}
& $\underline{55.64}${\tiny$\pm$0.36}
& $\mathbf{5.414}${\tiny$\pm$0.011}
& $\mathbf{0.444}${\tiny$\pm$0.001}
& $\underline{3.070}${\tiny$\pm$0.005} \\

HLR-LLM (Ours) & \textbf{V}+\textbf{M}+\textbf{B}
& $\mathbf{80.51}${\tiny$\pm$0.38}
& $\mathbf{76.19}${\tiny$\pm$0.11}
& $\mathbf{67.23}${\tiny$\pm$0.37}
& $\mathbf{80.87}${\tiny$\pm$0.20}
& $\mathbf{61.50}${\tiny$\pm$0.51}
& $4.461${\tiny$\pm$0.004}
& $\underline{0.439}${\tiny$\pm$0.001}
& $\mathbf{3.109}${\tiny$\pm$0.004} \\

\bottomrule
\end{tabular}%
}
\label{tab:model_full}
\vspace{-1mm}
\end{table}

A similar trend appears in \Tref{tab:model_full}.
Zero-shot models perform reasonably well on visibility, which can often be inferred from visual evidence, but struggle with tasks requiring spatial alignment and relational reasoning.
Fine-tuned baselines improve substantially, highlighting the importance of task-specific learning for spatial audio-visual understanding.
HLR-LLM achieves the best performance across all spatial relational reasoning tasks and consistently outperforms the dual-channel Qwen2.5-Omni baseline, further supporting the benefit of explicit binaural modeling for scene-level reasoning.
Compared with Hear You Are LLM, HLR-LLM achieves consistent gains across relational reasoning tasks, demonstrating stronger scene-level reasoning beyond source localization.
For scene captioning, HLR-LLM achieves the best SPIDEr score, while dual-channel Qwen2.5-Omni obtains the best CIDEr-D and METEOR scores.
Overall, the results show that explicit spatial audio modeling benefits not only source localization but also reasoning about sound sources in the visual scene.

\begin{table}[t]
\centering
\small
\renewcommand{\arraystretch}{0.90}
\caption{\textbf{Ablation study.}
We ablate input modalities and training strategies to validate the contribution of each component.
All metrics are reported in accuracy (\%).
\textbf{V}, \textbf{M}, and \textbf{B} denote visual input, monaural audio, and binaural audio, respectively.
Results are mean $\pm$ standard deviation over three seeds.
\textbf{Best} and \uline{second-best} mean results are shown in bold and underlined, respectively.
The full ablation is provided in Appendix~\ref{app:result}.
}
\vspace{-1mm}
\resizebox{1\linewidth}{!}{%
\begin{tabular}{@{}c l c cc cc ccc@{}}
\toprule
\multirow{2}{*}{\textbf{ID}}
& \multirow{2}{*}{\textbf{Variant}}
& \multirow{2}{*}{\textbf{Modality}}
& \multicolumn{2}{c}{\textbf{Egocentric Grounding}}
& \multicolumn{2}{c}{\textbf{Allocentric Grounding}}
& \multicolumn{3}{c}{\textbf{Spatial Relational Reasoning}} \\
\cmidrule(lr){4-5}\cmidrule(lr){6-7}\cmidrule(lr){8-10}
&
&
& Direction
& Distance
& Direction
& Distance
& Correspondence
& Hidden source
& Proximity \\
\midrule

\multicolumn{10}{l}{\textbf{\textit{Modality ablation}}} \\
\midrule

(a) & Without vision & \textbf{M}+\textbf{B}
& $74.72\pm0.36$
& $50.64\pm0.54$
& $21.51\pm0.41$
& $26.63\pm0.62$
& $57.11\pm0.11$
& $66.62\pm0.21$
& $34.13\pm0.12$ \\

(b) & Visual only & \textbf{V}
& $31.20\pm0.14$
& $20.33\pm0.49$
& $28.11\pm0.36$
& $22.96\pm0.22$
& $26.71\pm0.02$
& $24.16\pm0.43$
& $37.09\pm0.55$ \\

(c) & Without binaural & \textbf{V}+\textbf{M}
& $32.69\pm0.26$
& $30.92\pm0.42$
& $26.26\pm0.18$
& $27.73\pm1.22$
& $55.73\pm0.07$
& $51.24\pm0.58$
& $47.11\pm0.27$ \\

\midrule
\multicolumn{10}{l}{\textbf{\textit{Training ablation}}} \\
\midrule

(d) & Stage-2 training & \textbf{V}+\textbf{M}+\textbf{B}
& $77.92\pm0.05$
& $52.44\pm0.37$
& $\underline{48.37}\pm0.21$
& $38.83\pm0.96$
& $67.69\pm0.11$
& $66.52\pm0.35$
& $56.23\pm0.43$ \\

(e) & Joint training & \textbf{V}+\textbf{M}+\textbf{B}
& $\underline{78.00}\pm0.15$
& $\underline{53.04}\pm0.17$
& $45.01\pm0.59$
& $\underline{39.25}\pm0.68$
& $\underline{72.71}\pm0.23$
& $\mathbf{69.78}\pm0.24$
& $\underline{56.92}\pm0.21$ \\

(f) & HLR-LLM (Ours) & \textbf{V}+\textbf{M}+\textbf{B}
& $\mathbf{78.21}\pm0.34$
& $\mathbf{55.56}\pm0.27$
& $\mathbf{52.03}\pm0.26$
& $\mathbf{45.08}\pm0.85$
& $\mathbf{76.19}\pm0.11$
& $\underline{67.23}\pm0.37$
& $\mathbf{61.50}\pm0.51$ \\

\bottomrule
\end{tabular}%
}
\label{tab:ablation_spatial}
\end{table}

\subsection{Ablation study}

\Tref{tab:ablation_spatial} examines the contribution of each modality and training strategy.
Removing vision (a) still retains egocentric direction and distance performance, since binaural audio provides direct spatial cues.
However, it substantially reduces allocentric grounding and proximity reasoning, where visual object locations are required.
Interestingly, the model without vision still performs relatively well on hidden-source reasoning, suggesting that binaural cues provide useful evidence for distinguishing non-visible source configurations.
The visual-only model (b) performs poorly across most tasks, showing that vision alone is insufficient for reliable sound-source localization, especially when sources are off-screen or hidden.
Removing binaural audio (c) also causes a large drop in grounding tasks, showing that monaural audio lacks the spatial cues needed for accurate localization.
At the same time, its relatively stronger proximity performance suggests that visual context and semantic audio can still support some object-level relational reasoning.

The training ablation shows how each stage contributes to spatial reasoning. Stage-2 training (d), which includes Stages 1--2 but omits CoT tuning, already provides strong performance. Joint training with CoT (e), which mixes Stage-2 instruction data with CoT-formatted data, improves egocentric grounding, correspondence, and hidden-source reasoning, but does not consistently improve all grounding tasks. In contrast, our final model (f), which applies CoT tuning after Stage-2 instruction tuning, performs best across all grounding metrics and most relational reasoning tasks, with clear gains in allocentric grounding, correspondence, and proximity. These gains are most evident on tasks requiring multi-step spatial reasoning, such as transforming viewpoints or comparing object--source relations. Overall, the ablation indicates that binaural audio supports spatial localization, vision supports object-level grounding, and final-stage CoT tuning improves complex spatial reasoning.

\begin{table}[]
    \centering
    \caption{\textbf{Real-world adaptation on STARSS23.}
``Pretrain'' denotes HLR-AVSceneQA pretraining before STARSS23 fine-tuning.
``Sound'' denotes sound event recognition; direction, distance, and visibility are egocentric.
All metrics are accuracy (\%).
Results are mean $\pm$ standard deviation over three seeds.
\textbf{Best} and \uline{second-best} mean results are shown in bold and underlined, respectively.}
    \resizebox{0.75\linewidth}{!}{%
\begin{tabular}{llccccc}
\toprule
\textbf{ID} & \textbf{Model} & \textbf{Pretrain}
& \textbf{Sound} & \textbf{Direction} & \textbf{Distance} & \textbf{Visibility} \\
\midrule

(a) & Hear You Are LLM~\cite{ryu2025hear} & --
& $38.66\pm0.16$
& $15.25\pm0.57$
& $\underline{34.87}\pm0.35$
& $62.35\pm0.07$ \\

(b) & Hear You Are LLM~\cite{ryu2025hear} & \checkmark
& $40.80\pm0.76$
& $\underline{16.15}\pm0.36$
& $26.93\pm0.45$
& $63.80\pm0.48$ \\

(c) & HLR-LLM & --
& $\underline{44.59}\pm0.13$
& $16.13\pm0.04$
& $28.11\pm0.04$
& $\underline{69.67}\pm0.18$ \\

(d) & HLR-LLM & \checkmark
& $\mathbf{53.93}\pm0.14$
& $\mathbf{29.92}\pm0.13$
& $\mathbf{35.13}\pm0.07$
& $\mathbf{71.68}\pm0.09$ \\

\bottomrule
\end{tabular}%
}
    \label{tab:starss}
\end{table}

\subsection{Real-world adaptation}
\Tref{tab:starss} evaluates real-world adaptation on STARSS23. We fine-tune each model on the STARSS23 training set and compare performance with and without HLR-AVSceneQA pretraining. 
Comparing (a) vs. (b) and (c) vs. (d), HLR-AVSceneQA pretraining improves most evaluated tasks, suggesting that it provides useful spatial priors for adapting to real spatial audio-visual recordings.
Among models with HLR-AVSceneQA pretraining, HLR-LLM (d) outperforms Hear You Are LLM (b) across all metrics, with particularly large gains in sound recognition and direction grounding, demonstrating stronger adaptation to real-world spatial audio-visual data.

\begin{figure}
    \centering
    \includegraphics[width=1\linewidth]{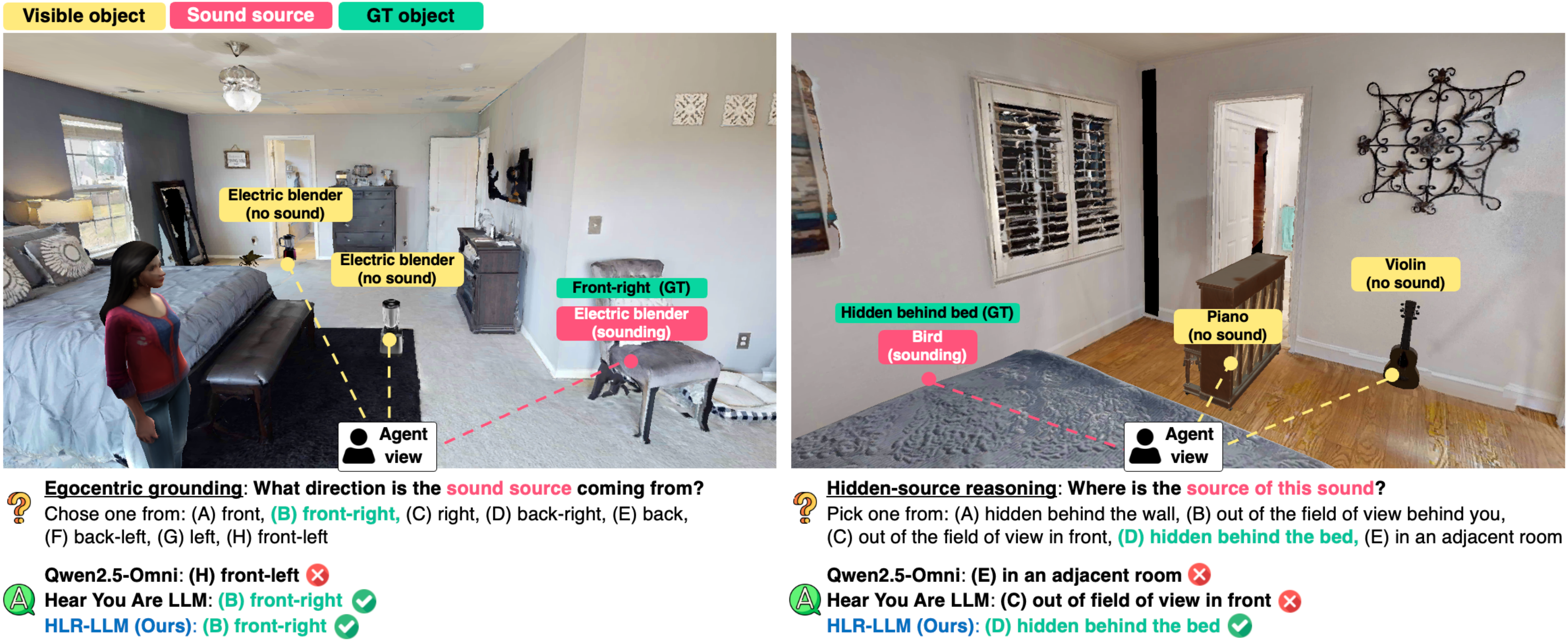}
    \vspace{-3mm}
    \caption{\textbf{Qualitative examples of spatial grounding.}
HLR-LLM correctly identifies the sound direction in a scene with multiple similar objects (left) and localizes a non-visible sound source as hidden behind the bed (right), outperforming the comparison models in the hidden-source example.}
    \label{fig:qual}
    
\end{figure}

\subsection{Qualitative results}
Figure~\ref{fig:qual} shows representative qualitative results for spatial grounding and relational reasoning.
In the left example, multiple visually similar electric blenders are present, while the sound originates from the front-right direction.
The monaural Qwen2.5-Omni predicts an incorrect direction, whereas models with binaural spatial input (Hear You Are LLM and HLR-LLM) correctly localize the source.
The right example is more challenging because the sounding bird is not visible and must be localized using spatial audio and scene layout.
While Qwen2.5-Omni and Hear You Are LLM incorrectly attribute the source to other non-visible regions, HLR-LLM correctly identifies it as hidden behind the bed.
Together, these examples show that explicit spatial audio cues help HLR-LLM localize sound sources and ground them within visible or hidden regions of the scene.
Further results and analyses across diverse spatial reasoning scenarios are provided in Appendix~\ref{app:result}.

%% file: sections/6_conclusion.tex
\section{Conclusion and discussion}\label{main:conclusion}
We introduce HLR-AVSceneQA, a comprehensive benchmark for spatial audio-visual scene understanding organized around a hear-localize-reason hierarchy. By covering spatial hearing, egocentric and allocentric grounding, and spatial relational reasoning, HLR-AVSceneQA evaluates whether AV-LLMs can move beyond recognizing audio-visual content to connect what is heard, where it originates, and how the source relates to the surrounding scene. We also propose HLR-LLM, which augments a strong AV-LLM with a binaural spatial audio branch and a staged curriculum with final-stage CoT supervision, improving spatial reasoning while retaining strong semantic audio-visual perception. Real-world adaptation experiments on STARSS23 further show that HLR-AVSceneQA pretraining provides useful spatial priors for downstream adaptation. These results highlight the importance of explicit spatial audio modeling and structured training for spatially aware AV-LLMs.

Our work also has limitations. HLR-AVSceneQA is built with a simulator-based pipeline, which provides controllable annotations but cannot fully capture the acoustic complexity and visual diversity of the real world. The benchmark focuses on indoor scenes, leaving dynamic agents, moving sound sources, and richer embodied interaction for future work. Future research should extend spatial audio-visual benchmarks to more realistic and dynamic settings and develop AV-LLMs that can actively perceive, move, and reason in spatial environments.

%% file: sections/7_appendix.tex
\clearpage
\renewcommand{\thefigure}{S\arabic{figure}}
\renewcommand{\thetable}{S\arabic{table}}
\setcounter{figure}{0} 
\setcounter{table}{0} 
\startcontents[appendix]

\section*{Table of Contents}
{
    \hypersetup{linkcolor=black}
    \printcontents[appendix]{}{1}{}
}

\section{Additional experimental details}\label{app:exp}

\subsection{Implementation details}
HLR-LLM is built on Qwen2.5-Omni with an additional binaural spatial audio pathway. For the spatial branch, we use a pretrained Spatial-AST encoder following BAT~\cite{zheng2024bat}. Spatial-AST produces audio tokens together with class tokens trained for semantic classification, direction prediction, and distance prediction. We use these tokens as spatial audio features and map them into the language embedding space with a lightweight projector. The projector is an MLP with a linear--GELU--linear structure, which has been shown to be effective for extending LLMs to multimodal inputs~\cite{liu2023visual,liu2024improved}.

Training follows the three-stage curriculum described in the main paper. In Stage~1, we train only the spatial alignment modules, including the spatial audio resampler and projector, while freezing the language model, visual encoder, monaural audio encoder, and spatial audio encoder. In Stages~2--3, we LoRA-finetune the language model and continue updating the spatial alignment modules while keeping the encoders frozen. We apply LoRA to all linear layers with rank 16, alpha 16, and dropout 0.05, and use LoRA+ with a larger learning-rate ratio for the LoRA $B$ matrix. We train with bfloat16 precision, FlashAttention, learning rate $2\times10^{-4}$, warmup ratio 0.05, weight decay 0.01, and gradient clipping at 1.0. Training is performed for 2 epochs on four NVIDIA A6000 GPUs with per-device batch size 4 and gradient accumulation 4.

\subsection{Details on comparison model experiments}
Since spatial AV-LLMs are still limited, we first evaluate existing AV-LLMs that accept visual input and monaural audio~\cite{yao2024minicpm,cheng2024videollama,tang2025video,xu2025qwen25omnitechnicalreport,xu2025qwen3,openai_realtime_api_docs,comanici2025gemini}. 
For these models, the binaural audio is converted to monaural during preprocessing. 
For SAVVY~\cite{chen2025savvy}, we follow the official code to process our inputs and evaluate only the applicable spatial grounding tasks, matching the scope of the original paper.

For fairer comparisons, we additionally fine-tune three baselines using the same direct-answer HLR-AVSceneQA training split.

\paragraph{Hear You Are LLM}
We fine-tune Hear You Are LLM~\cite{ryu2025hear}, the closest existing trainable model that uses both visual input and binaural audio for spatial audio-visual reasoning.
Although it was originally trained with panoramic observations, HLR-AVSceneQA is defined under normal-field-of-view (normal-FOV) inputs, where sources may be visible, occluded, off-screen, or located in another room.
We therefore adapt Hear You Are LLM to the same 1280$\times$720 normal-FOV images, 32-kHz binaural audio, QA prompts, answer targets, and evaluation protocol used for HLR-LLM.
Its SigLIP image encoder can directly process the normal-FOV images as patch tokens without architectural modification.
During fine-tuning, we apply LoRA to the image encoder (rank 16, alpha 32) and update the projectors/adapters and LLM, while keeping the audio encoder frozen.
We retain the original Hear You Are architecture and do not add the spatial audio branch used in HLR-LLM.
Accordingly, this comparison evaluates two adapted spatial AV-LLMs under matched data and evaluation conditions, rather than providing a fully architecture-matched comparison.

\paragraph{Qwen2.5-Omni}
We fine-tune Qwen2.5-Omni~\cite{xu2025qwen25omnitechnicalreport}, our monaural-audio base model, on HLR-AVSceneQA. 
This provides a controlled baseline for evaluating spatial reasoning without explicit binaural spatial modeling.

\paragraph{Dual-channel Qwen2.5-Omni}
We additionally fine-tune a dual-channel variant of Qwen2.5-Omni to test whether simply providing both binaural channels is sufficient for spatial reasoning. 
Instead of downmixing the binaural input to mono, the left and right channels are separately encoded through the original monaural audio pathway, and both resulting audio token streams are provided to the LLM. 
Unlike HLR-LLM, this variant does not introduce a dedicated spatial audio encoder or spatial projector. 
Thus, it isolates the benefit of explicit spatial-audio modeling from simply exposing the base model to both audio channels.

\subsection{STARSS23 preprocessing and QA construction}
\label{app:starss23}

STARSS23~\cite{shimada2023starss23} is an audio-visual dataset of real-world spatial recordings with synchronized video, first-order ambisonic (FOA) audio, and frame-level spatio-temporal annotations of sound events. 
We use STARSS23 as a real-world benchmark complementary to our simulator-generated HLR-AVSceneQA.
Starting from the official STARSS23 development split, we follow the official train/test partition and divide each recording into non-overlapping 10-second clips. 
Clips without a corresponding video stream are removed, resulting in 1,317 train clips and 1,116 test clips. 
For each clip, we extract a rectilinear egocentric image from the midpoint video frame. 
We also convert the FOA audio into binaural audio using the CIPIC subject-048 HRTF convention~\cite{algazi2001cipic}. 
This makes the real-world recordings compatible with the binaural audio interface used by our spatial audio-visual models while retaining the original STARSS23 annotations.

We convert STARSS23 azimuth annotations into our egocentric convention, where $0^\circ$ is front, $90^\circ$ is right, $180^\circ$ is back, and $270^\circ$ is left. 
We construct QA pairs for sound-event recognition, egocentric direction prediction, egocentric distance prediction, and visibility prediction. 
We distinguish static and dynamic sources using angular and distance variation of active annotation frames, and use only static sources for spatial QA. 
Following psychoacoustic studies of auditory motion and distance discrimination~\cite{saberi1990minimum,kolarik2016auditory}, a source is treated as static when its angular variation is within $5^\circ$ and its relative distance variation is within $20\%$.

For each retained static source, we compute representative azimuth and distance values from its active annotation frames and discretize them into the same egocentric bins used in HLR-AVSceneQA. 
We remove ambiguous QA items, including cases where frame-level annotations cross bin boundaries or multiple same-class sources make the queried source unclear. 
Since STARSS23 does not provide object-level semantic masks, visibility is defined geometrically by whether the representative source direction falls inside the canonical rectilinear camera field of view.

\paragraph{Scope of real-world tasks}
We restrict the real-world evaluation to tasks that can be reliably derived from STARSS23 annotations. 
STARSS23 does not provide the dense object- and scene-level annotations required for HLR-style hidden-source or allocentric grounding, including object geometry, occlusion state, room layout, and reference-object information. 
We therefore avoid constructing uncertain pseudo-labels for these unsupported tasks.

\begin{table}[t]
\centering
\small
\caption{\textbf{Ablation study on spatial hearing and spatial grounding tasks.}
Sound event recognition and grounding tasks are reported in accuracy (\%), while speech recognition is reported in WER (\%, lower is better).
\textbf{V}, \textbf{M}, and \textbf{B} denote visual input, monaural audio, and binaural audio, respectively.
Results are mean $\pm$ standard deviation over three seeds.
\textbf{Best} and \uline{second-best} mean results are shown in bold and underlined, respectively.}
\resizebox{\linewidth}{!}{%
\begin{tabular}{@{\,\,}c@{\,\,}l@{\,\,}c@{\,\,\,}cc@{\,\,\,}cc@{\,\,\,}cc@{\,\,}}
\toprule
\multirow{2}{*}{\textbf{ID}}
& \multirow{2}{*}{\textbf{Variant}}
& \multirow{2}{*}{\textbf{Modality}}
& \multicolumn{2}{c}{\textbf{Spatial Hearing}}
& \multicolumn{2}{c}{\textbf{Egocentric Grounding}}
& \multicolumn{2}{c}{\textbf{Allocentric Grounding}} \\
\cmidrule(lr){4-5}
\cmidrule(lr){6-7}
\cmidrule(lr){8-9}
& & &
Sound (Acc. \%)
& Speech (WER $\downarrow$)
& Direction
& Distance
& Direction
& Distance \\
\midrule

\multicolumn{9}{l}{\textbf{\textit{Modality ablation}}} \\
\midrule

(a) & Without vision & \textbf{M}+\textbf{B}
& $78.44\pm0.11$
& $14.62\pm2.81$
& $74.72\pm0.36$
& $50.64\pm0.54$
& $21.51\pm0.41$
& $26.63\pm0.62$ \\

(b) & Visual only & \textbf{V}
& $29.16\pm0.04$
& $250.14\pm15.37$
& $31.20\pm0.14$
& $20.33\pm0.49$
& $28.11\pm0.36$
& $22.96\pm0.22$ \\

(c) & Without binaural & \textbf{V}+\textbf{M}
& $80.57\pm0.27$
& $15.33\pm2.33$
& $32.69\pm0.26$
& $30.92\pm0.42$
& $26.26\pm0.18$
& $27.73\pm1.22$ \\

(d) & Without monaural & \textbf{V}+\textbf{B}
& $80.81\pm0.03$
& $361.05\pm4.00$
& $77.87\pm0.14$
& $47.78\pm0.49$
& $\underline{50.23}\pm0.41$
& $37.95\pm0.24$ \\

\midrule
\multicolumn{9}{l}{\textbf{\textit{Data scale ablation (Stage-2 training)}}} \\
\midrule

(e) & 10\% data & \textbf{V}+\textbf{M}+\textbf{B}
& $75.77\pm0.26$
& $\mathbf{10.65}\pm0.18$
& $75.39\pm0.07$
& $48.15\pm0.15$
& $29.31\pm0.11$
& $29.10\pm0.71$ \\

(f) & 50\% data & \textbf{V}+\textbf{M}+\textbf{B}
& $\mathbf{81.09}\pm0.13$
& $13.97\pm0.98$
& $76.22\pm0.05$
& $\mathbf{55.88}\pm0.19$
& $49.73\pm0.36$
& $\underline{41.20}\pm0.33$ \\

(g) & 100\% data & \textbf{V}+\textbf{M}+\textbf{B}
& $\mathbf{81.09}\pm0.28$
& $\underline{11.06}\pm0.11$
& $77.92\pm0.05$
& $52.44\pm0.37$
& $48.37\pm0.21$
& $38.83\pm0.96$ \\

\midrule
\multicolumn{9}{l}{\textbf{\textit{Training ablation (100\% data; compare (g)--(i))}}} \\
\midrule

(h) & Joint training & \textbf{V}+\textbf{M}+\textbf{B}
& $79.09\pm0.03$
& $12.19\pm0.43$
& $\underline{78.00}\pm0.15$
& $53.04\pm0.17$
& $45.01\pm0.59$
& $39.25\pm0.68$ \\

(i) & HLR-LLM (Ours) & \textbf{V}+\textbf{M}+\textbf{B}
& $\underline{81.03}\pm0.11$
& $14.43\pm0.54$
& $\mathbf{78.21}\pm0.34$
& $\underline{55.56}\pm0.27$
& $\mathbf{52.03}\pm0.26$
& $\mathbf{45.08}\pm0.85$ \\

\bottomrule
\end{tabular}%
}
\label{tab:ablation_hearing_grounding}
\end{table}

\begin{table}[t]
\centering
\small
\caption{\textbf{Ablation study on spatial relational reasoning and scene captioning tasks.}
Relational reasoning tasks are reported in accuracy (\%).
Scene captioning is evaluated with CIDEr-D, METEOR, and SPIDEr.
\textbf{V}, \textbf{M}, and \textbf{B} denote visual input, monaural audio, and binaural audio, respectively.
Results are mean $\pm$ standard deviation over three seeds.
\textbf{Best} and \uline{second-best} mean results are shown in bold and underlined, respectively.}
\resizebox{\linewidth}{!}{%
\begin{tabular}{@{\,\,}c@{\,\,}l@{\,\,}c@{\,\,\,}ccccc@{\,\,\,}ccc@{\,\,}}
\toprule
\multirow{2}{*}{\textbf{ID}}
& \multirow{2}{*}{\textbf{Variant}}
& \multirow{2}{*}{\textbf{Modality}}
& \multicolumn{5}{c}{\textbf{Spatial Relational Reasoning (Acc. \%)}}
& \multicolumn{3}{c}{\textbf{Scene Captioning}} \\
\cmidrule(lr){4-8}
\cmidrule(lr){9-11}
& & &
Visibility
& Correspondence
& Hidden source
& Loudness
& Proximity
& CIDEr-D
& METEOR
& SPIDEr \\
\midrule

\multicolumn{11}{l}{\textbf{\textit{Modality ablation}}} \\
\midrule

(a) & Without vision & \textbf{M}+\textbf{B}
& $65.43\pm0.42$
& $57.11\pm0.11$
& $66.62\pm0.21$
& $69.37\pm0.14$
& $34.13\pm0.12$
& $5.390\pm0.014$
& $0.441\pm0.001$
& $3.070\pm0.007$ \\

(b) & Visual only & \textbf{V}
& $53.21\pm0.10$
& $26.71\pm0.02$
& $24.16\pm0.43$
& $52.97\pm0.34$
& $37.09\pm0.55$
& $1.757\pm0.006$
& $0.316\pm0.001$
& $1.171\pm0.002$ \\

(c) & Without binaural & \textbf{V}+\textbf{M}
& $62.72\pm0.10$
& $55.73\pm0.07$
& $51.24\pm0.58$
& $68.84\pm0.32$
& $47.11\pm0.27$
& $4.796\pm0.014$
& $0.420\pm0.000$
& $2.751\pm0.007$ \\

(d) & Without monaural & \textbf{V}+\textbf{B}
& $75.67\pm0.07$
& $66.58\pm0.09$
& $66.48\pm0.10$
& $77.64\pm0.14$
& $55.23\pm0.62$
& $5.475\pm0.016$
& $0.440\pm0.001$
& $3.091\pm0.008$ \\

\midrule
\multicolumn{11}{l}{\textbf{\textit{Data scale ablation (Stage-2 training)}}} \\
\midrule

(e) & 10\% data & \textbf{V}+\textbf{M}+\textbf{B}
& $71.92\pm0.29$
& $65.72\pm0.24$
& $63.53\pm0.44$
& $74.75\pm0.22$
& $46.42\pm0.66$
& $4.946\pm0.013$
& $0.424\pm0.001$
& $2.842\pm0.008$ \\

(f) & 50\% data & \textbf{V}+\textbf{M}+\textbf{B}
& $76.74\pm0.20$
& $66.82\pm0.16$
& $66.17\pm0.48$
& $78.27\pm0.14$
& $55.54\pm0.60$
& $\underline{5.612}\pm0.017$
& $\underline{0.448}\pm0.001$
& $\underline{3.168}\pm0.009$ \\

(g) & 100\% data & \textbf{V}+\textbf{M}+\textbf{B}
& $77.62\pm0.07$
& $67.69\pm0.11$
& $66.52\pm0.35$
& $78.49\pm0.36$
& $56.23\pm0.43$
& $\mathbf{5.664}\pm0.010$
& $\mathbf{0.450}\pm0.000$
& $\mathbf{3.208}\pm0.005$ \\

\midrule
\multicolumn{11}{l}{\textbf{\textit{Training ablation (100\% data; compare (g)--(i))}}} \\
\midrule

(h) & Joint training & \textbf{V}+\textbf{M}+\textbf{B}
& $\underline{79.68}\pm0.12$
& $\underline{72.71}\pm0.23$
& $\mathbf{69.78}\pm0.24$
& $\underline{79.36}\pm0.21$
& $\underline{56.92}\pm0.21$
& $5.485\pm0.000$
& $0.439\pm0.001$
& $3.100\pm0.001$ \\

(i) & HLR-LLM (Ours) & \textbf{V}+\textbf{M}+\textbf{B}
& $\mathbf{80.51}\pm0.38$
& $\mathbf{76.19}\pm0.11$
& $\uline{67.23}\pm0.37$
& $\mathbf{80.87}\pm0.20$
& $\mathbf{61.50}\pm0.51$
& $4.461\pm0.004$
& $0.439\pm0.001$
& $3.109\pm0.004$ \\

\bottomrule
\end{tabular}%
}
\label{tab:ablation_reasoning_captioning}
\end{table}

\section{Additional results and analysis}\label{app:result}

\subsection{Ablation study}
Tables~\ref{tab:ablation_hearing_grounding} and~\ref{tab:ablation_reasoning_captioning} provide the full ablation results on HLR-AVSceneQA.
The modality ablation shows that binaural audio is essential for spatial grounding.
Removing the binaural pathway in (c) substantially degrades egocentric and allocentric grounding as well as spatial relational reasoning. The visual-only variant (b) also performs poorly across these tasks.
In contrast, removing monaural audio in (d) has a smaller effect on spatial grounding, but severely degrades far-field speech recognition, highlighting the role of the monaural pathway in preserving semantic audio understanding.

The data-scale ablation in (e)--(g) further disentangles the effect of training data from explicit spatial audio modeling.
Even with only 10\% of the training data, (e) substantially outperforms the 100\%-data model without binaural audio (c) on egocentric grounding and most relational reasoning tasks.
Increasing the training data generally improves spatial reasoning, particularly relational reasoning and scene captioning, although some grounding tasks exhibit non-monotonic variation.

The training ablation compares Stage-2 training (g), joint training (h), and our final staged curriculum with CoT tuning (i).
Final-stage CoT tuning improves both egocentric and allocentric grounding and achieves the strongest performance on most relational reasoning tasks, with particularly clear gains in correspondence and proximity.
The CoT supervision explicitly exposes intermediate steps, such as identifying the sound event, estimating the source location, transforming viewpoints, and comparing source--object relations, which helps the model decompose complex spatial queries before predicting the final answer.
We also observe that CoT tuning degrades scene captioning metrics and far-field ASR, suggesting a trade-off between structured multi-step reasoning and tasks that primarily require direct semantic prediction.
Overall, we select (i), which uses all modalities and final-stage CoT tuning, as our final model.

\begin{table}[t]
\centering
\small
\caption{\textbf{Zero-shot evaluation on STARSS23.}
We evaluate direct transfer to real-world spatial audio-visual recordings without any STARSS23 fine-tuning.
Sound denotes sound event recognition; direction, distance, and visibility are egocentric.
All metrics are reported in accuracy (\%).}
\resizebox{0.6\linewidth}{!}{%
\begin{tabular}{@{}lcccc@{}}
\toprule
\textbf{Model}
& \textbf{Sound}
& \textbf{Direction}
& \textbf{Distance}
& \textbf{Visibility} \\
\midrule


VideoLLaMA2~\cite{cheng2024videollama}
& $34.52\pm0.31$ & $12.61\pm0.16$ & $4.91\pm0.17$ & $61.80\pm0.31$ \\

Video-SALMONN 2~\cite{tang2025video}
& $35.30\pm0.11$ & $14.36\pm0.10$ & $4.80\pm0.29$ & $66.99\pm0.00$ \\

Qwen2.5-Omni~\cite{xu2025qwen25omnitechnicalreport}
& $\mathbf{43.22}\pm0.79$
& $14.66\pm0.76$
& $3.62\pm0.17$
& $67.61\pm0.25$ \\

HLR-LLM (Ours)
& $33.99\pm0.34$
& $\mathbf{17.48}\pm1.17$
& $\mathbf{19.76}\pm3.42$
& $\mathbf{68.61}\pm0.09$ \\

\bottomrule
\end{tabular}%
}
\label{tab:starss_zeroshot}
\end{table}

\subsection{Analysis on the STARSS23 dataset}
\Tref{tab:starss_zeroshot} evaluates direct transfer from HLR-AVSceneQA to real-world STARSS23 recordings without target-domain fine-tuning.
Despite being trained only on simulated scenes, HLR-LLM outperforms existing AV-LLMs on direction, distance, and visibility prediction, with particularly notable gains in distance estimation.
Although sound recognition remains less competitive, these results suggest that spatial grounding capabilities learned from HLR-AVSceneQA can transfer to real-world recordings without target-domain supervision.

\begin{figure}[t]
    \centering
    \includegraphics[width=1\linewidth]{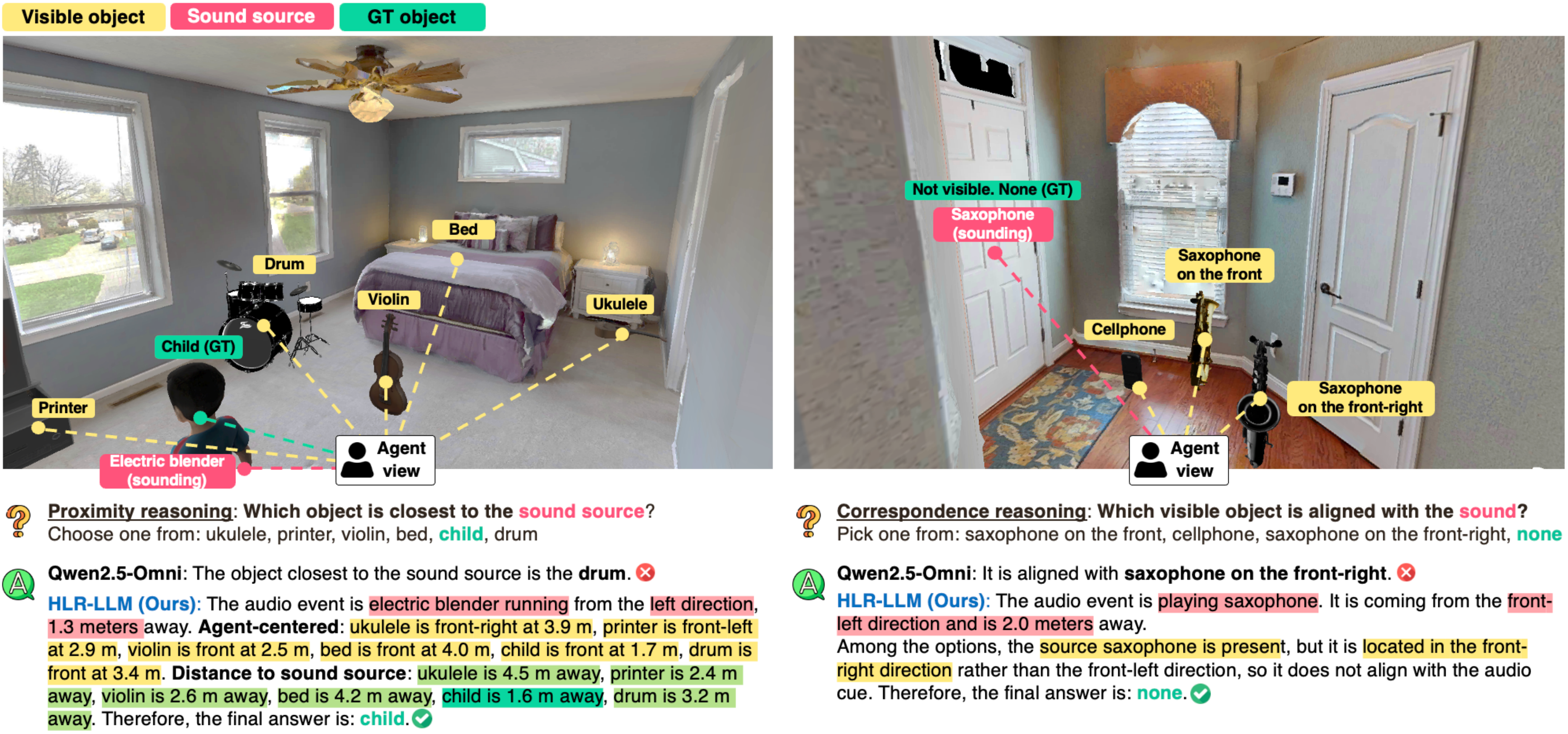}
    \vspace{-2mm}
    \caption{\textbf{Qualitative chain-of-thought reasoning examples from HLR-AVSceneQA.}
We compare Qwen2.5-Omni with HLR-LLM on proximity and audio-visual correspondence tasks.
HLR-LLM generates explicit CoT rationales that localize the sound source, relate it to visible objects, and derive the correct answer.}
    \label{fig:qualitative}
\end{figure}

\begin{figure}[t]
    \centering
    \includegraphics[width=1\linewidth]{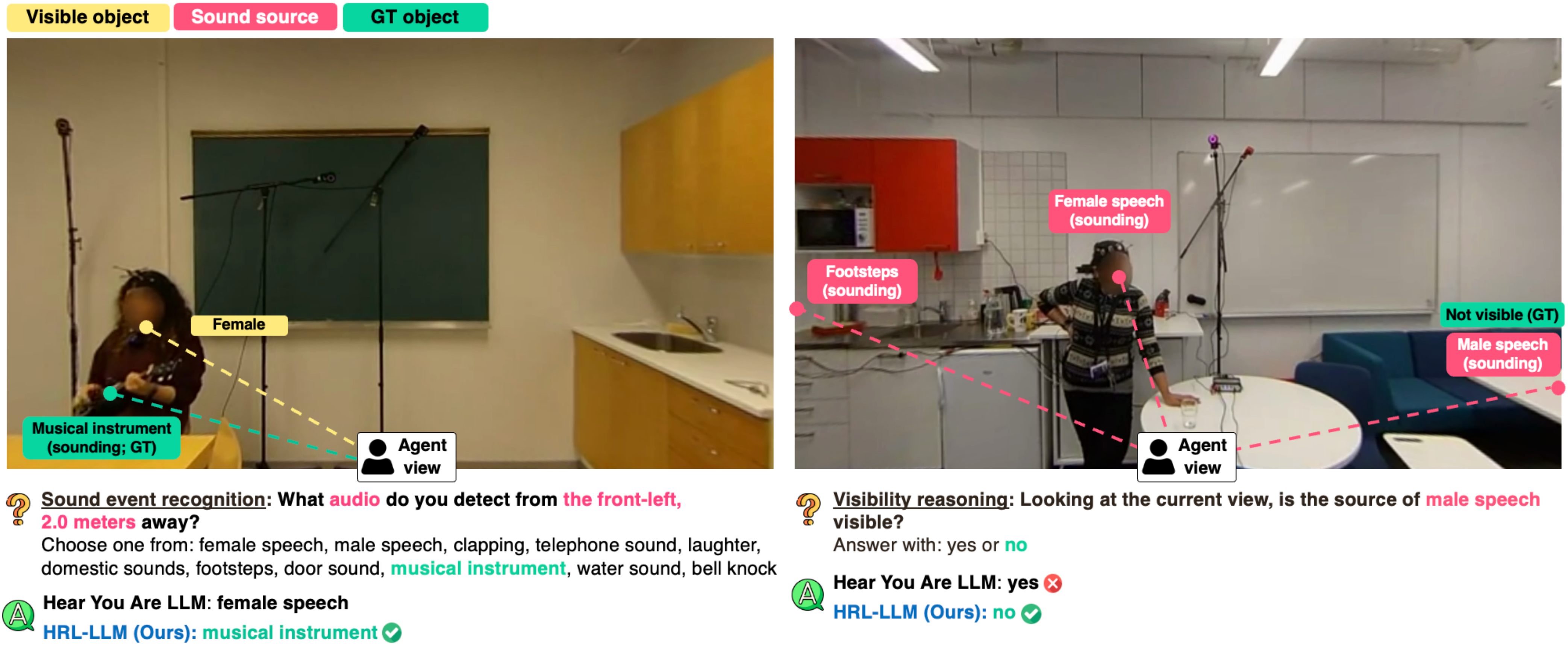}
    \vspace{-2mm}
    \caption{
    \textbf{Qualitative examples from the STARSS23 real-world evaluation.}
    We show two representative task types, sound-event recognition and visibility prediction, together with the midpoint video frame, annotated sound-source locations, ground-truth answers, and model predictions.}
    \label{fig:starss23_examples}
\end{figure}

\subsection{Qualitative results}
We present qualitative chain-of-thought (CoT) examples from HLR-AVSceneQA in Figure~\ref{fig:qualitative}.
The examples focus on spatial relational reasoning, where HLR-LLM explicitly decomposes the problem into sound-source localization and subsequent reasoning over visible objects.
In the proximity example, Qwen2.5-Omni selects the drum, whereas HLR-LLM's CoT first estimates the sound-source location and candidate-object positions, compares their distances to the source, and correctly identifies the child as the closest object.
In the correspondence example, Qwen2.5-Omni incorrectly selects a visible saxophone.
HLR-LLM's CoT instead localizes the sounding saxophone to the front-left, compares this position with the visible candidates, and correctly concludes that none of them is spatially aligned with the sound.
These examples illustrate how CoT supervision enables HLR-LLM to explicitly connect sound-source localization with downstream scene-level relational reasoning.

Figure~\ref{fig:starss23_examples} shows representative STARSS23 examples for sound-event recognition and visibility prediction.
HLR-LLM correctly recognizes the target event in multi-source audio and determines source visibility by combining spatial audio with visual evidence. These examples further demonstrate that the learned spatial audio-visual reasoning transfers to real-world recordings.

\begin{figure}[t]
    \centering
    \includegraphics[width=1\linewidth]{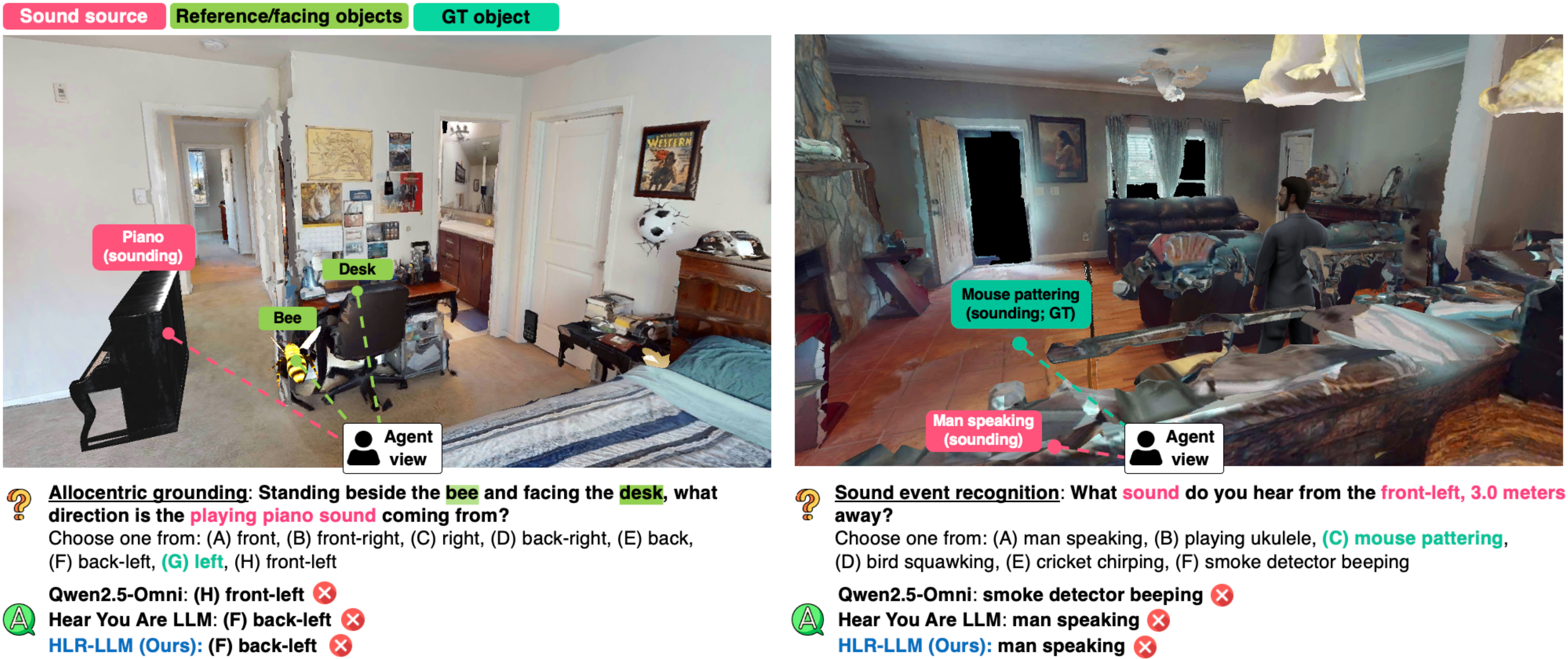}
    \vspace{-2mm}
    \caption{\textbf{Failure cases.}
We show representative failure cases where models make fine-grained spatial errors, such as confusing back-left with left, or struggle with multi-source scenes in which sound sources are spatially close.}
    \label{fig:failure_case}
\end{figure}

\begin{figure}[t]
    \centering
    \includegraphics[width=1\linewidth]{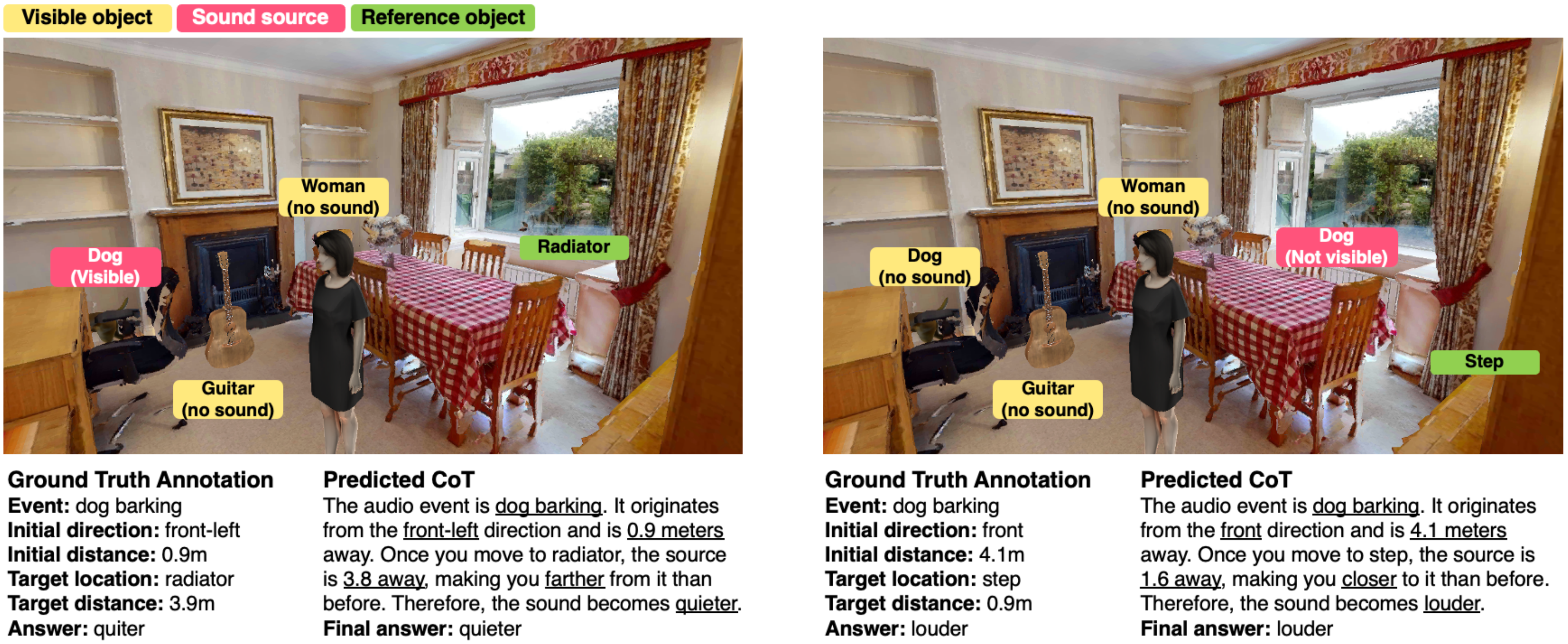}
    \vspace{-2mm}
    \caption{\textbf{Counterfactual analysis with conflicting audio-visual cues.} The image is kept fixed with the dog visible at its original location, while the binaural source position is changed. HLR-LLM follows the counterfactual spatial audio and correspondingly updates its localization and downstream loudness reasoning.}
    \label{fig:counterfactual}
\end{figure}

\paragraph{Failure cases}
We also provide failure cases in Fig.~\ref{fig:failure_case}. Most errors are caused by fine-grained spatial ambiguity. For example, the model may confuse adjacent direction labels, such as \texttt{back-right} vs. \texttt{right} or \texttt{front-left} vs. \texttt{left}, especially when the source lies near a bin boundary. Errors also occur when several candidate objects are spatially clustered, making it difficult to associate the localized sound with the correct object. In multi-source scenes, failures often arise when two sources are close to each other and the non-target source is louder. In such cases, overlapping acoustic cues can dominate the prediction, causing the model to localize or reason about the wrong source.

\paragraph{Counterfactual analysis}
We further conduct a counterfactual analysis to examine whether HLR-LLM relies on spatial audio cues when visual and auditory evidence conflict.
We keep the rendered image fixed, where the dog remains visible at its original location, and replace the binaural audio with dog barking rendered from different spatial positions.
As shown in Fig.~\ref{fig:counterfactual}, HLR-LLM follows the counterfactual spatial audio in both cases.
In the left case, it localizes the barking to the front-left at 0.9\,m and infers that moving to the radiator increases the source distance, resulting in a quieter prediction.
In the right case, it instead localizes the barking to the front at 4.1\,m and predicts that moving to the step brings the agent closer to the source, resulting in a louder prediction, even though the visible dog remains at its original front-left location.
These results provide evidence that HLR-LLM uses spatial audio for localization and subsequent spatial reasoning rather than relying solely on the visible source position.

\begin{figure}[t]
    \centering
    \includegraphics[width=1\linewidth]{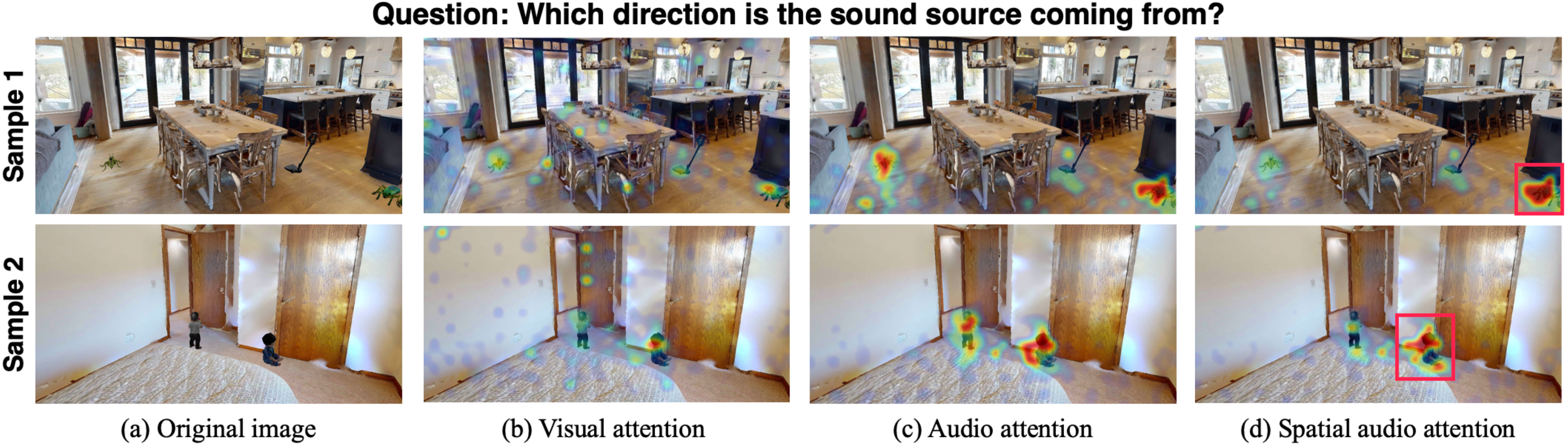}
    \vspace{-2mm}
    \caption{\textbf{Modality-specific attention visualization.}
    From left to right: input image, visual-to-visual, audio-to-visual, and spatial-audio-to-visual attention from the 18th transformer layer.
    Spatial-audio tokens attend more selectively to the ground-truth sound-source regions (red boxes).}
    \label{fig:attention}
\end{figure}

\paragraph{Modality-specific attention analysis}
To examine how modality-specific representations interact with visual content, we extract self-attention weights from the 18th transformer layer of HLR-LLM.
For each modality, we use its corresponding tokens as queries and the visual tokens as keys, averaging the attention weights across all query tokens and attention heads.
We visualize three modality-to-visual attention maps: visual-to-visual using $Z_{\mathrm{vis}}$, audio-to-visual using $Z_{\mathrm{aud}}$, and spatial-audio-to-visual using $Z_{\mathrm{spa}}$.
The visual-to-visual map reflects regions emphasized by the visual representation, while the latter two indicate which image regions are attended to by the conventional audio and spatial-audio representations, respectively.
As shown in Fig.~\ref{fig:attention}, spatial-audio tokens attend more selectively to the ground-truth sound-source regions than conventional audio tokens, providing qualitative evidence that the spatial pathway captures localization-relevant visual associations.

\section{Details on HLR-AVSceneQA}\label{app:detail}
\label{app:spatialav_sceneqa_details}

We provide implementation-level details on HLR-AVSceneQA, including the dataset construction pipeline, task characteristics, question and answer formats, quality-control procedures, and additional details that were not introduced in the main paper.

\subsection{Scene initialization, agent placement, and scene generation}
\label{app:scene_initialization_agent_placement}

\paragraph{Simulator setup}
We generate each sample in Habitat-Sim with acoustic rendering using the SoundSpaces simulator~\cite{chen2022soundspaces}. The agent is initialized with synchronized RGB, semantic, depth, panoramic RGB, and binaural audio sensors. The RGB sensor is a pinhole camera used as the main visual input. The semantic and depth sensors are used for visibility checking and geometric supervision, while the panoramic RGB sensor is used for visualization and debugging. The binaural audio sensor is co-located with the visual agent and uses a two-channel receiver at a 32 kHz sampling rate with material-aware indirect propagation. We set the sensor height to 150 cm and the camera field of view to $90^\circ$.

We select valid rooms from HM3D scenes~\cite{yadav2023habitat} by checking that the semantic region contains a floor, a valid room-level region of interest, and sufficient navigable positions. In the standard setting, we place the agent near a room corner and orient it toward the room center, which increases the chance that inserted objects and room semantics are visible. Since this placement may bias sound sources toward the front view, we also sample agent poses closer to the room center. This makes it possible to construct scenes in which one reference object remains visible while another object or sound source falls outside the camera view.

\paragraph{Coordinate system and azimuth definition}
We use Habitat coordinates for camera, agent, and source placement, and convert positions into metric distances when constructing QA labels. For azimuth, we annotate angles around the horizontal plane, ranging from 0 to $360^\circ$. In our setup, $0^\circ$ indicates the front direction, $90^\circ$ indicates the right direction, $180^\circ$ indicates the back direction, and $270^\circ$ indicates the left direction relative to the agent. We do not consider elevation in this manuscript, as human spatial-audio localization is generally known to be more accurate and robust in azimuth than in elevation~\cite{ege2018accuracy}.

For both direction and distance, we discretize continuous spatial values into bins to make
training and evaluation more stable. For azimuth, we use 8 bins, each covering $45^\circ$.
Angles increase clockwise from straight ahead: \texttt{front} corresponds to $337.5^\circ$--$22.5^\circ$,
\texttt{right} to $67.5^\circ$--$112.5^\circ$, \texttt{back} to $157.5^\circ$--$202.5^\circ$, and \texttt{left} to $247.5^\circ$--$292.5^\circ$. For distance, we use 10 bins, with a 1-meter interval for 0–1 m, 0.5-meter intervals from 1 to 5 m, and a final bin for distances beyond 5 m.
The full bin definitions are shown in \Tref{tab:azimuth_distance_bins}.
This discretization reduces ambiguity from small geometric differences and facilitates stable training and evaluation.
\begin{table}[t]
    \centering
    \small
    \caption{\textbf{Azimuth and distance bins used in HLR-AVSceneQA.}
    Azimuth angles increase clockwise from the agent's forward direction.}
    \label{tab:azimuth_distance_bins}

    \begin{minipage}[c]{0.48\linewidth}
        \centering
        \begin{tabular}{lll}
        \toprule
        \textbf{Type} & \textbf{Bin} & \textbf{Range} \\
        \midrule
        \multirow{8}{*}{Azimuth}
        & front       & $337.5^\circ$--$22.5^\circ$ \\
        & front-right & $22.5^\circ$--$67.5^\circ$ \\
        & right       & $67.5^\circ$--$112.5^\circ$ \\
        & back-right  & $112.5^\circ$--$157.5^\circ$ \\
        & back        & $157.5^\circ$--$202.5^\circ$ \\
        & back-left   & $202.5^\circ$--$247.5^\circ$ \\
        & left        & $247.5^\circ$--$292.5^\circ$ \\
        & front-left  & $292.5^\circ$--$337.5^\circ$ \\
        \midrule
        \multirow{10}{*}{Distance}
        & \texttt{A} & $0$--$1.0$ m \\
        & \texttt{B} & $1.0$--$1.5$ m \\
        & \texttt{C} & $1.5$--$2.0$ m \\
        & \texttt{D} & $2.0$--$2.5$ m \\
        & \texttt{E} & $2.5$--$3.0$ m \\
        & \texttt{F} & $3.0$--$3.5$ m \\
        & \texttt{G} & $3.5$--$4.0$ m \\
        & \texttt{H} & $4.0$--$4.5$ m \\
        & \texttt{I} & $4.5$--$5.0$ m \\
        & \texttt{J} & $>5.0$ m \\
        \bottomrule
        \end{tabular}
    \end{minipage}
    \hfill
    \begin{minipage}[c]{0.48\linewidth}
        \centering
        \includegraphics[width=\linewidth]{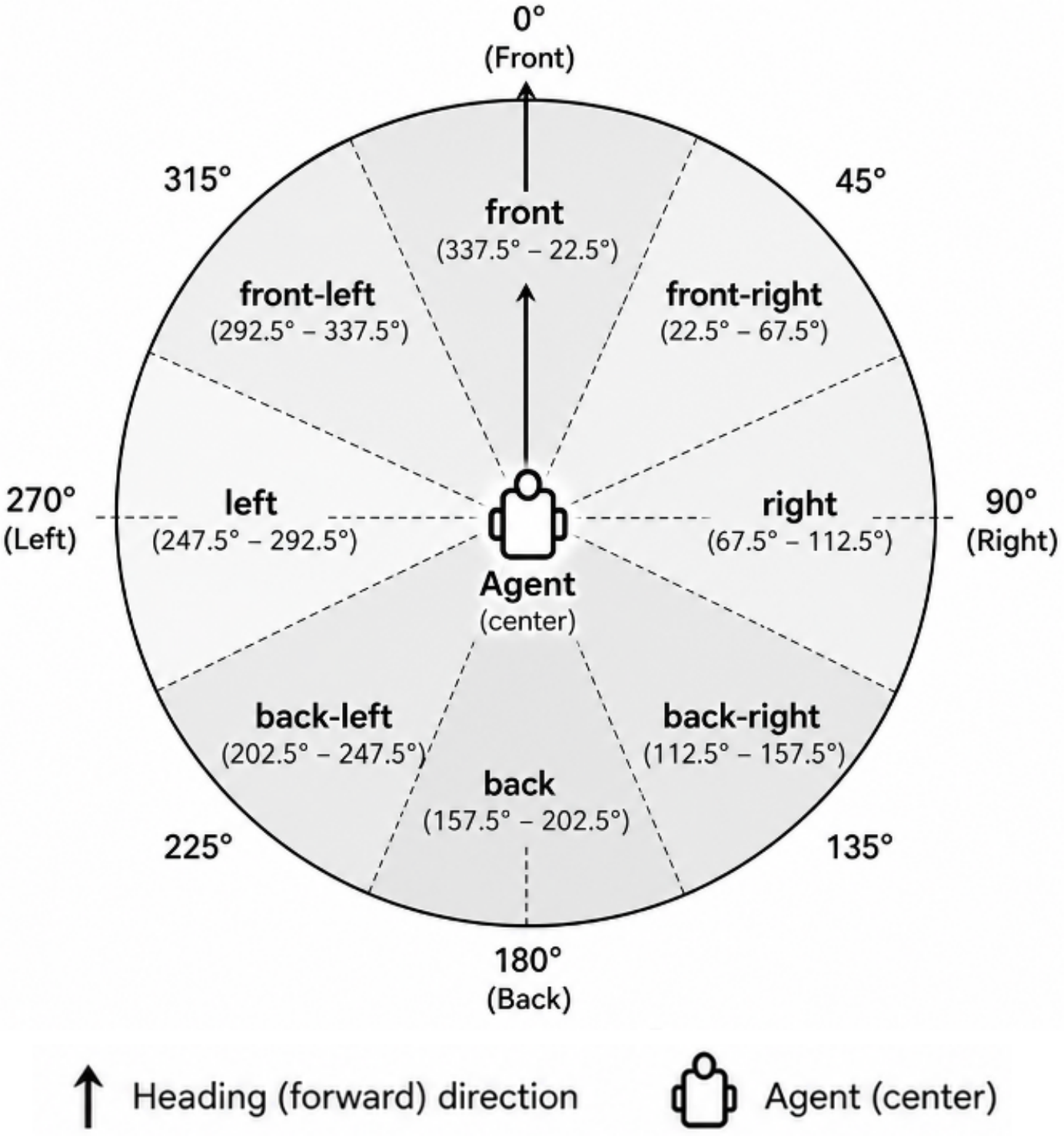}
    \end{minipage}
\end{table}

\begin{figure}[t]
    \centering
    \includegraphics[width=1\linewidth]{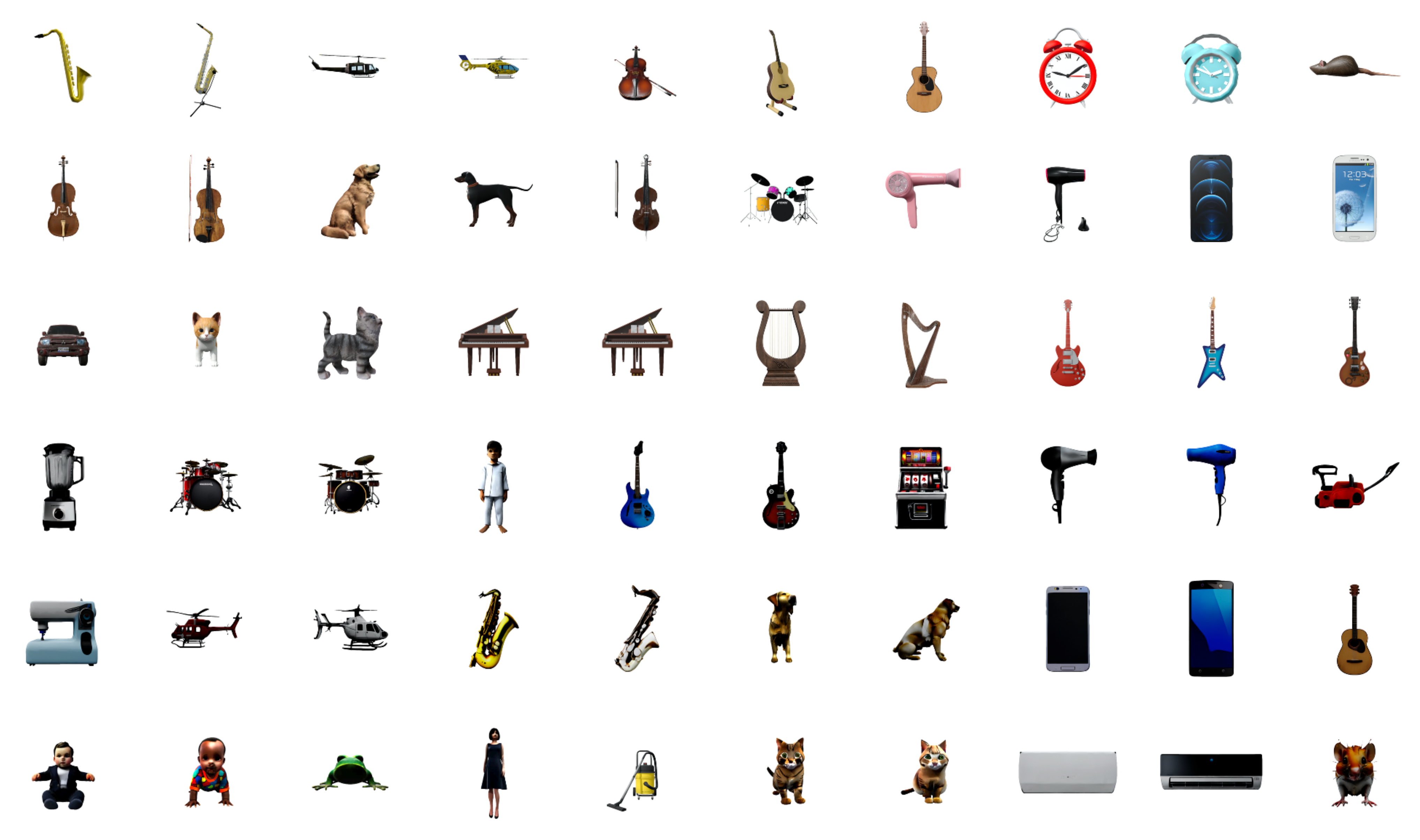}
    \caption{\textbf{Examples of 3D object meshes.} The first three rows show meshes retrieved from Objaverse-XL, while the last three rows show meshes generated by generative models.}
    \label{fig:appendix_mesh}
\end{figure}

\paragraph{3D object meshes and scene insertion}
We insert 3D object meshes into HM3D rooms to synthesize controlled layouts containing both sounding and non-sounding objects. The object pool is built from curated meshes, including Objaverse-XL assets~\cite{deitke2023objaverse}, as well as generated meshes for sound categories whose object coverage is insufficient. For generated sounding-object meshes, we first use Stable Diffusion 2~\cite{rombach2022high} for text-to-image synthesis, and then use TRELLIS~\cite{xiang2025structured} for image-to-3D mesh generation. The size and brightness of each object category are manually determined to ensure plausible rendering in indoor scenes. Examples are shown in \Fref{fig:appendix_mesh}.

When inserting objects into a room, we reject placements that collide with room geometry, overlap existing objects above a horizontal IoU threshold, or violate minimum agent-to-object or object-to-object distance constraints. These filtering steps help ensure that inserted objects are physically plausible and visually distinguishable.

\begin{figure}[t]
    \centering
    \includegraphics[width=1\linewidth]{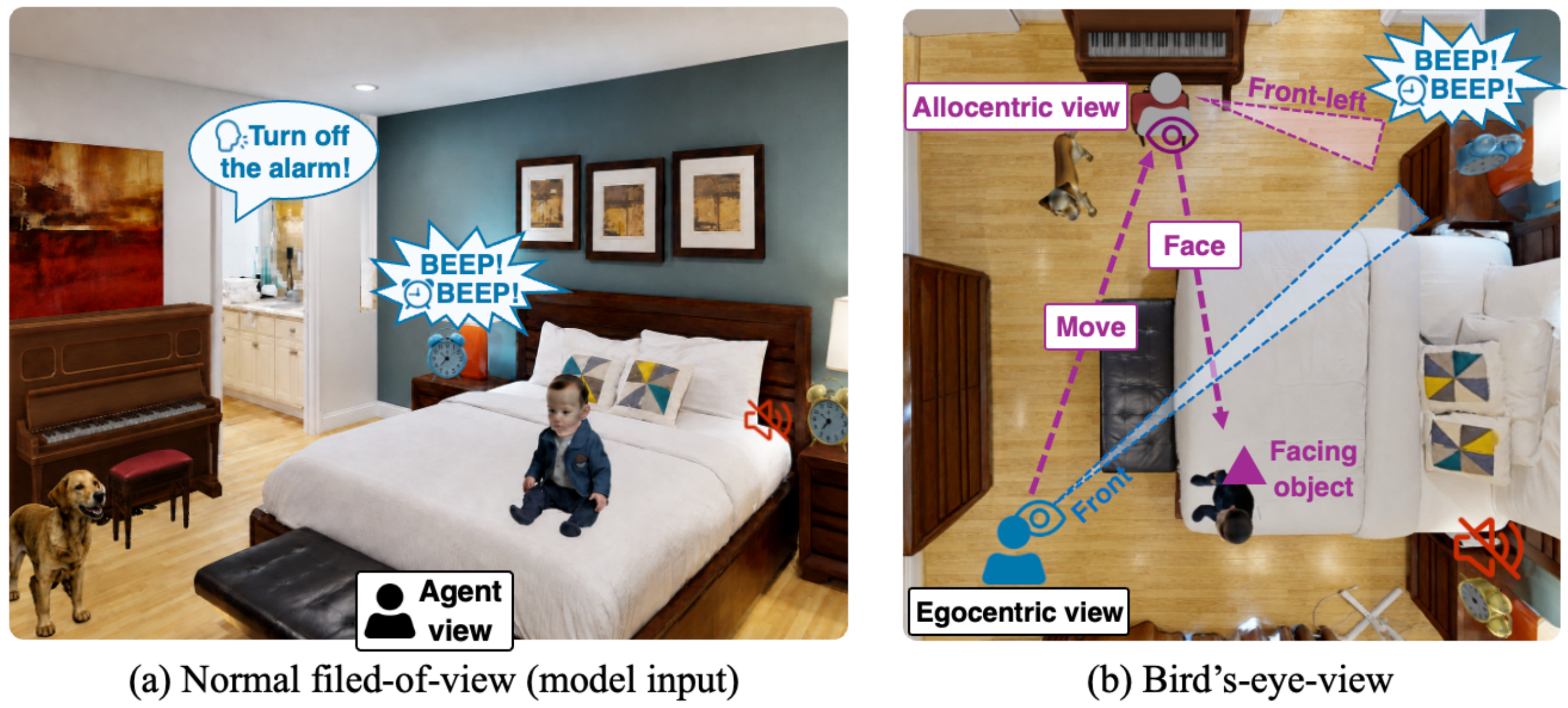}
    \caption{\textbf{Example illustration.} This example data sample is used to facilitate understanding of our task definitions and details.}
    \label{fig:appendix_example}
\end{figure}

\paragraph{Binaural audio generation}
For each active source, we render binaural audio using SoundSpaces 2.0. Let $x_j[t]$ be the dry monaural waveform assigned to source $j$, located at $\mathbf{s}_j$, and let the agent receiver be at position $\mathbf{r}$ with heading $\theta$. SoundSpaces produces a channel-wise binaural room impulse response:
\begin{equation}
    h_{m,j}[\tau] = \mathcal{R}_{m}(\tau; \mathbf{s}_j,\mathbf{r},\theta,\mathcal{E}),
    \qquad m \in \{L,R\},
\end{equation}
where $\mathcal{E}$ denotes the scene geometry and material configuration. The binaural signal is then obtained by convolving the normalized dry waveform $\tilde{x}_j$ with the corresponding room impulse response:
\begin{equation}
    y_{m,j}[t] = (h_{m,j} * \tilde{x}_j)[t]
    = \sum_{\tau=0}^{T_h-1} h_{m,j}[\tau] \tilde{x}_j[t-\tau].
\end{equation}
For multiple active sources, the rendered signals are summed:
\begin{equation}
    y_m[t] = \sum_{j=1}^{N} y_{m,j}[t], \qquad m \in \{L,R\}.
\end{equation}
The final stereo waveform $\mathbf{y}[t]=[y_L[t],y_R[t]]$ is saved at a 32 kHz sampling rate.

\paragraph{Spatial reasoning scenarios}
We intentionally include several scenarios that require audio-visual spatial reasoning beyond semantic matching. \textit{Occluded sound sources} place the active source behind a visible object or behind a wall, so the source may project into the camera view while having no visible semantic pixels. \textit{Ambiguous sound sources among same-category instances} include multiple objects of the same category, requiring the model to determine which instance is spatially aligned with the sound. \textit{Semantically mismatched but spatially aligned cases} attach a sound event to an object that does not semantically match the event, preventing models from relying only on category correspondence. \textit{Adjacent-room sound sources} move the source into a neighboring room connected through the scene geometry, creating cases where the source is not visible but can still be heard from a distant location.

\subsection{Task definitions and details}
\label{app:task_definitions_details}
Here we provide additional details on the task definitions, their descriptions, and the intention of each task, using the example in \Fref{fig:appendix_example}. In this scenario, \Fref{fig:appendix_example}~(a) shows the agent's view, which is the model input with a normal field of view. Several instances are inserted, including a dog, a piano, a baby, and two alarm clocks. There are two sound events: one is a man speaking, saying ``Turn off the alarm,'' coming from the agent's front-left direction and originating from an adjacent room. The other is an alarm ringing from the blue alarm clock, coming from the front direction, 3.4 meters away. Based on this example, we describe the goal and intended evaluation behavior of each task.

Furthermore, to improve linguistic diversity and reduce overfitting to fixed question formats, we generate paraphrased templates for each task. We create approximately 15 paraphrases per task and show three representative prompts below. The full set of templates will be included in the dataset release.

\paragraph{Spatial hearing: sound event recognition}
This task evaluates semantic audio perception. The model is asked to identify the audible sound event from a set of candidate sound categories. A total of 10 candidate sound categories are provided as options. The target sound category is always included, while distractors are sampled from semantically related categories, sound categories associated with scene objects, and random categories. When multiple sources are present, the prompt may additionally specify the source direction and distance to indicate which source should be recognized.
\begin{itemize}
    \item \textit{\textbf{Q:} What sound do you hear from the front, 3.4 meters away?}
    \item \textit{\textbf{A:} alarm clock ringing.}
    \item \textit{\textbf{Paraphrase 1:} Which sound is coming from the front, 3.4 meters away?}
    \item \textit{\textbf{Paraphrase 2:} Identify the sound event from the front direction at 3.4 meters.}
\end{itemize}

\paragraph{Spatial hearing: far-field automatic speech recognition}
This task evaluates far-field automatic speech recognition (ASR) from spatialized audio. Since speech may contain important information that the agent should understand, the model is required to recognize speech as part of spatial hearing. Because speech can originate from both nearby and distant locations, far-field ASR is more challenging than clean or near-field ASR. We ask the model to return only the exact quoted transcription, without any additional explanation.
\begin{itemize}
    \item \textit{\textbf{Q:} Transcribe what the person said. Output only the transcription inside quotes.}
    \item \textit{\textbf{A:} ``Turn off the alarm.''}
    \item \textit{\textbf{Paraphrase 1:} Write the exact phrase spoken. Return only the quoted transcription.}
    \item \textit{\textbf{Paraphrase 2:} Provide the exact spoken words inside quotes.}
\end{itemize}

\paragraph{Spatial grounding: egocentric direction prediction}
This task evaluates egocentric sound-source direction estimation. The answer is computed from the vector pointing from the agent to the sound source, with its azimuth measured relative to the agent's current forward direction. We ask the model to answer using a discretized 8-bin azimuth label as shown in \Tref{tab:azimuth_distance_bins}. If multiple sources are present, we specify the queried sound source.
\begin{itemize}
    \item \textit{\textbf{Q:} From this camera perspective, which direction bin does the source of alarm clock ringing fall into?}
    \item \textit{\textbf{A:} front.}
    \item \textit{\textbf{Paraphrase 1:} Based on the current view, which direction is the alarm clock ringing coming from?}
    \item \textit{\textbf{Paraphrase 2:} From this viewpoint, identify the direction bin for the source of alarm clock ringing.}
\end{itemize}

\paragraph{Spatial grounding: egocentric distance prediction}
This task evaluates egocentric source-distance estimation. The answer is computed as the horizontal distance between the agent and the active sound source, and is then converted into a discrete distance bin, as shown in \Tref{tab:azimuth_distance_bins}. If multiple sources are present, we specify the queried sound source.
\begin{itemize}
    \item \textit{\textbf{Q:} From this camera perspective, what is the distance to the source of alarm clock ringing?}
    \item \textit{\textbf{A:} F.}
    \item \textit{\textbf{Paraphrase 1:} Based on the current view, how far away is the source of alarm clock ringing?}
    \item \textit{\textbf{Paraphrase 2:} From this viewpoint, select the distance bin for the source of alarm clock ringing.}
\end{itemize}

\paragraph{Spatial grounding: allocentric direction prediction}
This task evaluates direction reasoning from a different viewpoint. The prompt asks the model to imagine standing at a reference object and facing another object, as shown in \Fref{fig:appendix_example}~(b). The source direction is then recomputed in this new coordinate frame. It is intended to evaluate whether the model can reason from a different perspective and understand the 3D spatial structure of the current scene. The task is generated only when valid visible reference and facing objects are available. We avoid prompts where the reference and facing objects have identical final category names, which would make the instruction ambiguous. As in the egocentric case, the answer is a discretized 8-bin azimuth label. If multiple sources are present, we specify the queried sound source.
\begin{itemize}
    \item \textit{\textbf{Q:} If you stand by the piano facing the baby, which direction is the source of alarm clock ringing coming from?}
    \item \textit{\textbf{A:} front-left.}
    \item \textit{\textbf{Paraphrase 1:} Standing next to the piano and facing the baby, which direction does the alarm clock ringing come from?}
    \item \textit{\textbf{Paraphrase 2:} From a position at the piano facing the baby, identify the direction bin of the alarm clock ringing.}
\end{itemize}

\paragraph{Spatial grounding: allocentric distance prediction}
This task evaluates distance reasoning from a different viewpoint. Similar to allocentric direction prediction, the prompt asks the model to imagine standing at a reference object and facing another object. The source distance is then recomputed from the reference object's position. This task evaluates whether the model can reason about source distance from a different location in the scene. We avoid prompts where the reference and facing objects have identical final category names, which would make the instruction ambiguous. As in the egocentric case, the answer is a discretized 10-bin distance label. If multiple sources are present, we specify the queried sound source.
\begin{itemize}
    \item \textit{\textbf{Q:} If you stand by the piano facing the baby, what is the distance to the source of alarm clock ringing?}
    \item \textit{\textbf{A:} B.}
    \item \textit{\textbf{Paraphrase 1:} Standing next to the piano and facing the baby, how far away is the alarm clock ringing?}
    \item \textit{\textbf{Paraphrase 2:} From a position at the piano facing the baby, select the distance bin for the source of alarm clock ringing.}
\end{itemize}

\paragraph{Spatial relational reasoning: visibility}
This task evaluates whether the model can determine if the sound source is visible from the current viewpoint. This task becomes difficult when the scene contains the same or similar categories, but the actual sound comes from outside the current viewpoint. For example, one may incorrectly assume that the man speaking sound in \Fref{fig:appendix_example} comes from the baby, since the baby is also human-related. However, with sufficient spatial audio perception, the model should distinguish that the baby is not making the sound and that the sound source is not visible. The ground truth is computed from semantic rendering: a source is considered visible if the number of semantic pixels corresponding to the source object exceeds a configured threshold. If multiple sources are present, we specify the queried sound source.
\begin{itemize}
    \item \textit{\textbf{Q:} From the current camera perspective, is the source of man speaking visible?}
    \item \textit{\textbf{A:} no.}
    \item \textit{\textbf{Paraphrase 1:} Based on the current view, can you see the source of man speaking?}
    \item \textit{\textbf{Paraphrase 2:} From this viewpoint, is the man speaking source in view?}
\end{itemize}

\paragraph{Spatial relational reasoning: correspondence}
This task evaluates audio-visual spatial correspondence. The model is asked to identify which visible object is spatially aligned with the sound source. The correct answer may be semantically different from the sound category, since the task focuses on spatial alignment rather than category matching. This task becomes challenging when multiple visible objects exist, as in \Fref{fig:appendix_example}. Without spatial audio understanding, the model may confuse whether the yellow alarm clock is ringing. If no visible object corresponds to the sound source, the answer is \texttt{None}. When multiple visible inserted objects share the same category, we disambiguate them using orientation labels, such as \textit{``alarm clock on the front-right.''} If multiple sources are present, we specify the queried sound source.
\begin{itemize}
    \item \textit{\textbf{Q:} From the current camera perspective, which object is spatially aligned with the sound of alarm clock ringing?}
    \item \textit{\textbf{A:} alarm clock in front.}
    \item \textit{\textbf{Paraphrase 1:} Which visible object corresponds to where the alarm clock ringing comes from?}
    \item \textit{\textbf{Paraphrase 2:} Which object best matches the spatial origin of the alarm clock ringing?}
\end{itemize}

\paragraph{Spatial relational reasoning: hidden-source localization}
This task evaluates reasoning about non-visible sound sources. It is asked when the source is not reliably visible in the current image. The answer includes five options: hidden behind the wall, hidden behind a specific object, in an adjacent room, out of the field of view in front, and out of the field of view behind the agent. This task goes one step beyond visibility prediction. The model must reason where the source may be if it is not visible. It should decide whether the sound comes from a distant place, such as another room, or whether it should be in the field of view but is occluded by an object, which requires accurate visual perception. The ground-truth label is determined using visibility, projection, adjacent-room metadata, and occluder information. If multiple sources are present, we specify the queried sound source.
\begin{itemize}
    \item \textit{\textbf{Q:} You hear a man speaking. Where is the source of this sound?}
    \item \textit{\textbf{A:} in an adjacent room.}
    \item \textit{\textbf{Paraphrase 1:} Given the sound of man speaking, where is its source?}
    \item \textit{\textbf{Paraphrase 2:} A man speaking is audible. Where could the source be situated?}
\end{itemize}

\paragraph{Spatial relational reasoning: loudness change}
This task evaluates relative acoustic reasoning under hypothetical movement. The prompt asks whether the sound would become louder or quieter if the agent moved to a specified object. Here, louder or quieter is defined as a controlled relative movement judgment based on whether the source becomes closer or farther, rather than an estimate of absolute perceived loudness. This assesses whether the model can decide whether a movement would bring the agent closer to or farther from the sound source. It also goes one step beyond allocentric distance prediction. The model must estimate the source distance from the changed location and compare it with the current source distance. The label is computed by comparing the current agent-to-source distance with the candidate-object-to-source distance. If the candidate object is closer to the source than the current agent position, the answer is \texttt{louder}; otherwise, it is \texttt{quieter}. If multiple sources are present, we specify the queried sound source.
\begin{itemize}
    \item \textit{\textbf{Q:} If you move to the piano, will the sound of man speaking become louder or quieter?}
    \item \textit{\textbf{A:} louder.}
    \item \textit{\textbf{Paraphrase 1:} When at the piano, will the sound of man speaking be louder or quieter?}
    \item \textit{\textbf{Paraphrase 2:} Upon moving to the piano, does the man speaking become louder or quieter?}
\end{itemize}

\paragraph{Spatial relational reasoning: proximity reasoning}
This task goes one step beyond loudness change and evaluates source-to-object spatial reasoning. The model is given a set of candidate objects and must select the one closest to the sound source. This requires comparing the source distance to each candidate object, similar to evaluating loudness change for each object, and then ranking the candidates to select the closest one. The ground truth is computed using the Euclidean distance from the source to each candidate object. As with other distance-based relational tasks, we exclude adjacent-room sources to avoid cases where direct Euclidean geometry does not reflect the effective acoustic propagation path. Tied cases are skipped to avoid ambiguous supervision. If multiple sources are present, we specify the queried sound source.
\begin{itemize}
    \item \textit{\textbf{Q:} Which object is closest to the source of alarm clock ringing?}
    \item \textit{\textbf{A:} bed.}
    \item \textit{\textbf{Paraphrase 1:} To which object is the alarm clock ringing nearest?}
    \item \textit{\textbf{Paraphrase 2:} Which object is at the shortest distance from the source of alarm clock ringing?}
\end{itemize}

\paragraph{Spatial reasoning: scene captioning}
This task evaluates compositional spatial audio-visual scene description. The model is asked to describe each sound source using complete sentences and include the sound event, direction, visibility, and hidden-source location when applicable. The prompt explicitly instructs the model not to describe unrelated scene content. When multiple sources are present, the model must provide a separate description for each source.
\begin{itemize}
    \item \textit{\textbf{Q:} Describe only the sound source or sound sources.}
    \item \textit{\textbf{A:} The sound event is alarm clock ringing. It is coming from the front direction. The sound source is visible in the scene.}
    \item \textit{\textbf{Paraphrase 1:} Summarize only the sound source information.}
    \item \textit{\textbf{Paraphrase 2:} For each source, state the sound event, direction, visibility, and hidden location if applicable.}
\end{itemize}

\subsection{Chain-of-thought supervision}
\label{app:cot_supervision}
We additionally construct chain-of-thought-style supervision to encourage the model to connect audio perception, visual grounding, and spatial reasoning before producing the final answer. 
Both the final answers and CoT rationales are deterministically derived from structured simulator metadata through task-specific templates rather than free-form generation. GPT-4~\cite{openai2023gpt4} is used only to paraphrase question wording for linguistic diversity.
This is especially important for tasks that cannot be solved from a single cue alone, such as allocentric reasoning, hidden-source localization, loudness change, and proximity reasoning. In these tasks, the model must first identify the sound event, estimate its direction or distance, relate it to visible or non-visible objects, and then convert this intermediate spatial evidence into the requested answer. We keep the reasoning brief and require the model to end with a normalized final-answer line.

\paragraph{Spatial grounding: allocentric direction prediction}
\begin{itemize}
    \item \textit{\textbf{Q:} If you are located at the cabinet facing the door frame, determine which direction the source of woman singing is coming from. Reason briefly, then end with exactly one line: ``Final answer: <8bin\_direction\_label>''.}
    \item \textit{\textbf{A:} The audio event woman singing is heard from the front direction at 1.2 m.\\
    The cabinet is positioned front-left at a distance of 2.6 m from the current location.\\
    After moving to cabinet while maintaining the same orientation, the sound source appears at right with a distance of 2.0 m.\\
    From this position, door frame is found in the front-right direction at 3.5 m.\\
    Rotate right to face door frame, establishing the new forward axis.\\
    In this new frame, the sound source lies front-right.\\
    Final answer: front-right}
\end{itemize}

\paragraph{Spatial grounding: allocentric distance prediction}
\begin{itemize}
    \item \textit{\textbf{Q:} Standing next to the cabinet and facing the typewriter, what is the distance to the sound source? Reason briefly, then end with exactly one line: ``Final answer: <distance\_bin>''.}
    \item \textit{\textbf{A:} The sound event is playing acoustic guitar, initially at 1.0 m in the right direction.\\
    The reference object cabinet is positioned front-right at a distance of 1.7 m.\\
    After moving to cabinet, the sound source is approximately 1.2 m away.\\
    This measurement places it within the 1-1.5 m interval.\\
    Final answer: B}
\end{itemize}

\paragraph{Spatial relational reasoning: visibility}
\begin{itemize}
    \item \textit{\textbf{Q:} From this perspective, is the sound source in sight? Answer with [yes, no]. Reason briefly, then end with exactly one line: ``Final answer: <yes\_or\_no>''.}
    \item \textit{\textbf{A:} The audio event is playing clarinet.\\
    It is coming from the back direction and is 1.0 meters away.\\
    In that direction, the clarinet is not present, so the sound source cannot be seen. Thus, the source is not visible.\\
    Final answer: no}
\end{itemize}

\paragraph{Spatial relational reasoning: correspondence}
\begin{itemize}
    \item \textit{\textbf{Q:} From the camera's perspective, which object is at the location of the sound of mouse squeaking? Pick one option; select None if no object in view is producing sound. Reason briefly, then end with exactly one line: ``Final answer: <object\_or\_None>''.}
    \item \textit{\textbf{A:} The audio event is mouse squeaking.\\
    It is coming from the left direction and is 1.6 meters away.\\
    Among the options, no visible source in the left direction matches the sound.\\
    Final answer: None}
\end{itemize}

\paragraph{Spatial relational reasoning: hidden-source localization}
\begin{itemize}
    \item \textit{\textbf{Q:} The sound of woman speaking is present. Where is it originating from? Choose one from: (A)--(E). Reason briefly, then end with exactly one line: ``Final answer: <hidden\_source\_option>''.}
    \item \textit{\textbf{A:} The audio event is woman speaking.\\
    Although it comes from the right direction, the source lies outside the current field of view in that direction.\\
    Therefore, the source is out of the field of view in front.\\
    Final answer: out of the field of view in front}
\end{itemize}

\paragraph{Spatial relational reasoning: loudness change}
\begin{itemize}
    \item \textit{\textbf{Q:} By moving to pillow, does the sound become louder or quieter? Choose one from: [louder, quieter]. Reason briefly, then end with exactly one line: ``Final answer: <louder\_or\_quieter>''.}
    \item \textit{\textbf{A:} The audio event is hair dryer drying.\\
    It originates from the front direction and is 4.0 meters away.\\
    Once you move to pillow, the source is 2.8 m away, making you closer to it than before.\\
    Therefore, the sound becomes louder.\\
    Final answer: louder}
\end{itemize}

\paragraph{Spatial relational reasoning: proximity reasoning}
\begin{itemize}
    \item \textit{\textbf{Q:} Which object is closest to the sound source? Select one from: [bed, vacuum cleaner, chair, air conditioning]. Reason briefly, then end with exactly one line: ``Final answer: <object\_label>''.}
    \item \textit{\textbf{A:} The audio event is man speaking.\\
    It originates from the back-right direction and is 0.7 meters away.\\
    Agent-centered candidate layout: bed is front-right at 1.6 m from the agent, vacuum cleaner is front-right at 3.0 m from the agent, chair is front at 3.4 m from the agent, and air conditioning is front at 2.0 m from the agent.\\
    Comparing source-to-candidate distances gives: bed is 1.4 m away from the sound source, vacuum cleaner is 3.1 m away from the sound source, chair is 3.7 m away from the sound source, and air conditioning is 2.4 m away from the sound source.\\
    Therefore, the nearest object is bed.\\
    Final answer: bed}
\end{itemize}

\subsection{Quality control}\label{app:quality_control}
\paragraph{Automatic quality control}
We apply several filtering and retry rules to avoid noisy supervision. 
First, rooms are skipped if they do not have a valid floor, contain too few insertion slots, or lack enough valid navigable points. Second, object placements are rejected if they collide with room geometry, overlap existing objects above the IoU threshold, violate minimum distance constraints, or fail semantic visibility checks. Third, generated 3D meshes and assets from Objaverse-XL are manually inspected by human annotators. We retain only sufficiently high-quality objects, filtering out low-quality meshes and implausible categories. We also find that, in some cases, sounding objects, reference objects, facing objects, or candidate objects used in task options can be too small to be clearly visible. We therefore apply a pixel-count threshold and remove QA pairs that ask about objects that are hardly visible. Finally, proximity questions with tied nearest objects are skipped. 

\paragraph{Human verification}
We additionally conduct a human verification study on 150 samples, covering 1,144 QA pairs, with three human annotators.
Annotators verify whether the target sound is audible, whether its approximate spatial location can be identified, whether candidate and reference objects mentioned in the questions are observable, and whether the annotated source visibility and hidden-source location are consistent with the audio-visual scene.
As shown in Table~\ref{tab:human_verification}, 1,133 of 1,144 QA pairs pass human verification, corresponding to an overall pass rate of 99.0\%.
All spatial-hearing and egocentric-grounding questions achieve a 100\% pass rate, while the remaining task categories achieve pass rates between 96.8\% and 100\%.
These results provide human validation that our metadata- and template-based construction pipeline produces reliable spatial audio-visual QA annotations across the benchmark tasks.

\begin{table}[t]
\centering
\scriptsize
\caption{\textbf{Human verification of HLR-AVSceneQA annotations.}
Three human annotators verify 1,144 QA pairs from 150 sampled scenes.
We report the number of passed and failed QA pairs and the corresponding pass rate for each task.}
\label{tab:human_verification}
\resizebox{0.75\linewidth}{!}{%
\begin{tabular}{llrrrr}
\toprule
\textbf{Task family}
& \textbf{Task}
& \textbf{Total}
& \textbf{Passed}
& \textbf{Failed}
& \textbf{Pass rate} \\
\midrule

\multirow{2}{*}{Spatial hearing}
& Sound event recognition
& 112 & 112 & 0 & 100.0\% \\
& Far-field ASR
& 29 & 29 & 0 & 100.0\% \\

\midrule
\multirow{2}{*}{Egocentric grounding}
& Direction
& 98 & 98 & 0 & 100.0\% \\
& Distance
& 98 & 98 & 0 & 100.0\% \\

\midrule
\multirow{3}{*}{Allocentric grounding}
& Direction
& 148 & 146 & 2 & 98.6\% \\
& Distance
& 148 & 148 & 0 & 100.0\% \\

\midrule
\multirow{5}{*}{Relational reasoning}
& Visibility
& 107 & 105 & 2 & 98.1\% \\
& Hidden source
& 66 & 65 & 1 & 98.5\% \\
& Loudness change
& 111 & 109 & 2 & 98.2\% \\
& Proximity
& 126 & 122 & 4 & 96.8\% \\
& Correspondence
& 101 & 101 & 0 & 100.0\% \\

\midrule
\textbf{Overall}
& --
& \textbf{1,144}
& \textbf{1,133}
& \textbf{11}
& \textbf{99.0\%} \\

\bottomrule
\end{tabular}%
}
\end{table}

\begin{table}[t]
\centering
\scriptsize
\caption{\textbf{Dataset statistics by task split.} 
Each entry reports the number of QA pairs. 
For Train and Test, Single and Multi are computed from the source-count annotations. 
For Train-CoT, Single and Multi report the number of CoT QA pairs in each source-count group after task-specific filtering.}
\label{tab:appendix_dataset_statistics}
\resizebox{\linewidth}{!}{%
\begin{tabular}{llrrr|rrr|rrr}
\toprule
\multirow{2}{*}{\textbf{Task family}} 
& \multirow{2}{*}{\textbf{Task}}
& \multicolumn{3}{c|}{\textbf{Train}}
& \multicolumn{3}{c|}{\textbf{Train-CoT}}
& \multicolumn{3}{c}{\textbf{Test}} \\
\cmidrule(lr){3-5}\cmidrule(lr){6-8}\cmidrule(lr){9-11}
& & \textbf{Single} & \textbf{Multi} & \textbf{Total}
  & \textbf{Single} & \textbf{Multi} & \textbf{Total}
  & \textbf{Single} & \textbf{Multi} & \textbf{Total} \\
\midrule

\multirow{2}{*}{Spatial hearing}
& Sound event 
& 14,794 & 7,054 & 21,848
& -- & -- & --
& 1,498 & 772 & 2,270 \\

& Far-field ASR 
& 2,708 & 867 & 3,575
& -- & -- & --
& 380 & 128 & 508 \\

\midrule
\multirow{4}{*}{Spatial grounding}
& Egocentric direction 
& 13,770 & 6,500 & 20,270
& -- & -- & --
& 1,442 & 632 & 2,074 \\

& Egocentric distance 
& 13,770 & 6,500 & 20,270
& -- & -- & --
& 1,442 & 632 & 2,074 \\

& Allocentric direction 
& 8,410 & 3,909 & 12,319
& 8,086 & 3,735 & 11,821
& 1,454 & 461 & 1,915 \\

& Allocentric distance 
& 8,410 & 3,909 & 12,319
& 8,086 & 3,735 & 11,821
& 1,454 & 461 & 1,915 \\

\midrule
\multirow{6}{*}{\makecell[l]{Spatial relational\\reasoning}}
& Visibility 
& 11,277 & 5,350 & 16,627
& 11,277 & 5,350 & 16,627
& 1,644 & 410 & 2,054 \\

& Correspondence 
& 14,794 & 7,054 & 21,848
& 14,794 & 7,054 & 21,848
& 2,066 & 268 & 2,334 \\

& Hidden-source 
& 5,173 & 2,693 & 7,866
& 5,173 & 1,321 & 6,494
& 1,094 & 382 & 1,476 \\

& Loudness change 
& 12,542 & 5,860 & 18,402
& 12,542 & 5,860 & 18,402
& 1,164 & 291 & 1,455 \\

& Proximity 
& 5,166 & 2,391 & 7,557
& 5,166 & 2,340 & 7,506
& 720 & 248 & 968 \\

& Scene captioning 
& 10,968 & 1,192 & 12,160
& -- & -- & --
& 2,581 & 184 & 2,765 \\

\midrule
\textbf{Total}
& --
& \textbf{121,782} & \textbf{53,279} & \textbf{175,061}
& \textbf{65,124} & \textbf{29,395} & \textbf{94,519}
& \textbf{16,939} & \textbf{4,869} & \textbf{21,808} \\
\bottomrule
\end{tabular}%
}
\end{table}

\begin{table}[t]
\centering
\scriptsize
\caption{\textbf{Additional dataset statistics and train--test overlap.}
We report label distributions for hidden-source and visibility tasks, followed by statistics on the train and test splits.
Percentages are computed within each corresponding task split.
Train and test environments, rooms, and object mesh instances are disjoint, while test semantic categories are covered by the training split.}
\label{tab:additional_dataset_statistics}
\resizebox{0.7\linewidth}{!}{%
\begin{tabular}{lccc}
\toprule
\textbf{Statistic}
& \textbf{Train}
& \textbf{Test}
& \textbf{Train--Test Overlap} \\
\midrule

\multicolumn{4}{l}{\textbf{\textit{Hidden-source distribution}}} \\
\midrule
Off-screen (front)
& 2,681 (34.1\%)
& 408 (27.6\%)
& -- \\

Off-screen (back)
& 2,752 (35.0\%)
& 518 (35.1\%)
& -- \\

Adjacent room
& 959 (12.2\%)
& 237 (16.1\%)
& -- \\

Hidden behind wall
& 418 (5.3\%)
& 123 (8.3\%)
& -- \\

Hidden behind object
& 1,056 (13.4\%)
& 190 (12.9\%)
& -- \\

\textbf{Total}
& \textbf{7,866}
& \textbf{1,476}
& -- \\

\midrule
\multicolumn{4}{l}{\textbf{\textit{Visibility distribution}}} \\
\midrule
Visible
& 8,912 (53.6\%)
& 631 (30.7\%)
& -- \\

Not visible
& 7,715 (46.4\%)
& 1,423 (69.3\%)
& -- \\

\textbf{Total}
& \textbf{16,627}
& \textbf{2,054}
& -- \\

\midrule
\multicolumn{4}{l}{\textbf{\textit{Split statistics and overlap}}} \\
\midrule
Environments
& 145
& 36
& 0 \\

Rooms
& 1,028
& 241
& 0 \\

Sound categories
& 108
& 79
& 79 \\

Object categories
& 37
& 37
& 37 \\

Object mesh instances
& 3,018
& 1,299
& 0 \\

\bottomrule
\end{tabular}%
}
\end{table}

\subsection{Dataset statistics}\label{app:dataset_statistics}
Table~\ref{tab:appendix_dataset_statistics} reports the number of generated QA pairs for the training and test splits, including the number of CoT supervision examples for each task and the breakdown of QA pairs constructed from single-source and multi-source scenes.
Table~\ref{tab:additional_dataset_statistics} further characterizes the benchmark composition by reporting the distributions of hidden-source and visibility cases, together with train--test overlap statistics.
The hidden-source test set covers diverse non-visible configurations, including off-screen, adjacent-room, wall-occluded, and object-occluded sources.
The visibility task contains 53.6\% visible and 46.4\% non-visible cases in training, and 30.7\% visible and 69.3\% non-visible cases in the test set.
Importantly, the training and test splits use disjoint environments and rooms, as well as disjoint object mesh instances, while all test sound and object categories are represented in the training split.
Thus, the benchmark evaluates generalization to unseen environments, scene layouts, and object instances while maintaining consistent semantic task coverage.

\Fref{fig:dist} and \Fref{fig:dir} further visualize the distributions of distance and direction bins for egocentric and allocentric settings in both the training and test sets.
Finally, \Fref{fig:insert} and \Fref{fig:audio} show the top 30 inserted object categories and active sound-source categories in the training and test sets.

\section{Broader impact}\label{app:broad}
HLR-AVSceneQA and HLR-LLM aim to support spatially aware audio-visual understanding for embodied AI, assistive agents, robotics, and accessibility applications. The ability to recognize what sound is present, localize where it originates, and reason about how it relates to the current scene is especially important for agents that navigate indoor environments and assist in daily situations. For example, such systems may help localize alarms, identify speakers, or reason about events outside the camera view. Our benchmark is simulator-based and does not contain private real-world recordings. However, future deployment of spatial audio-visual systems should follow strict privacy, consent, and data-minimization practices, especially when operating in real homes or shared indoor spaces.

\section{LLM usage}\label{app:llm}
\begin{figure}[t]
    \centering
    \includegraphics[width=1\linewidth]{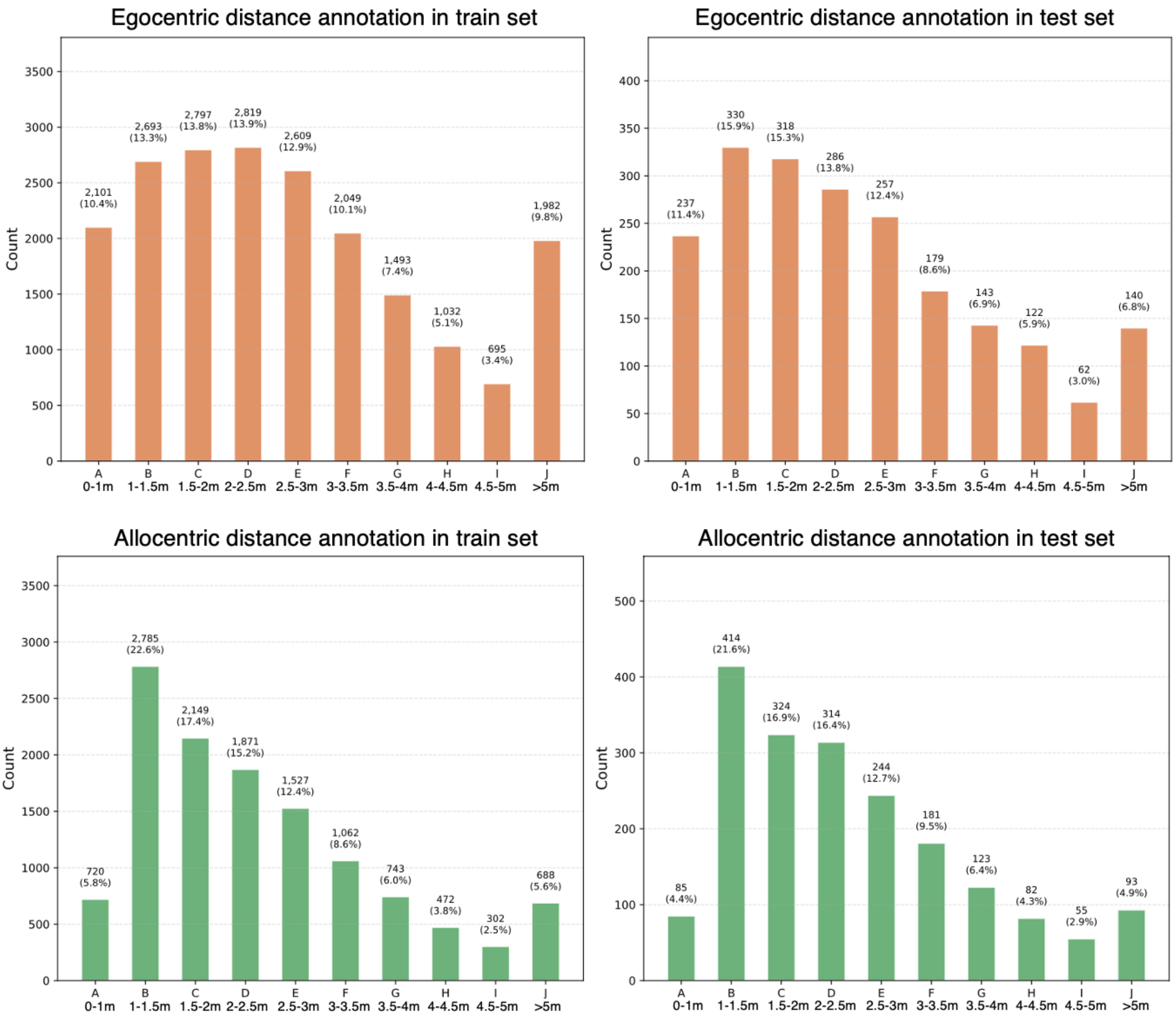}
    \caption{\textbf{Distance annotation statistics.} We show histograms of distance-bin annotations for egocentric and allocentric perspectives in both the training and test sets.}
    \label{fig:dist}
\end{figure}

\begin{figure}[t]
    \centering
    \includegraphics[width=1\linewidth]{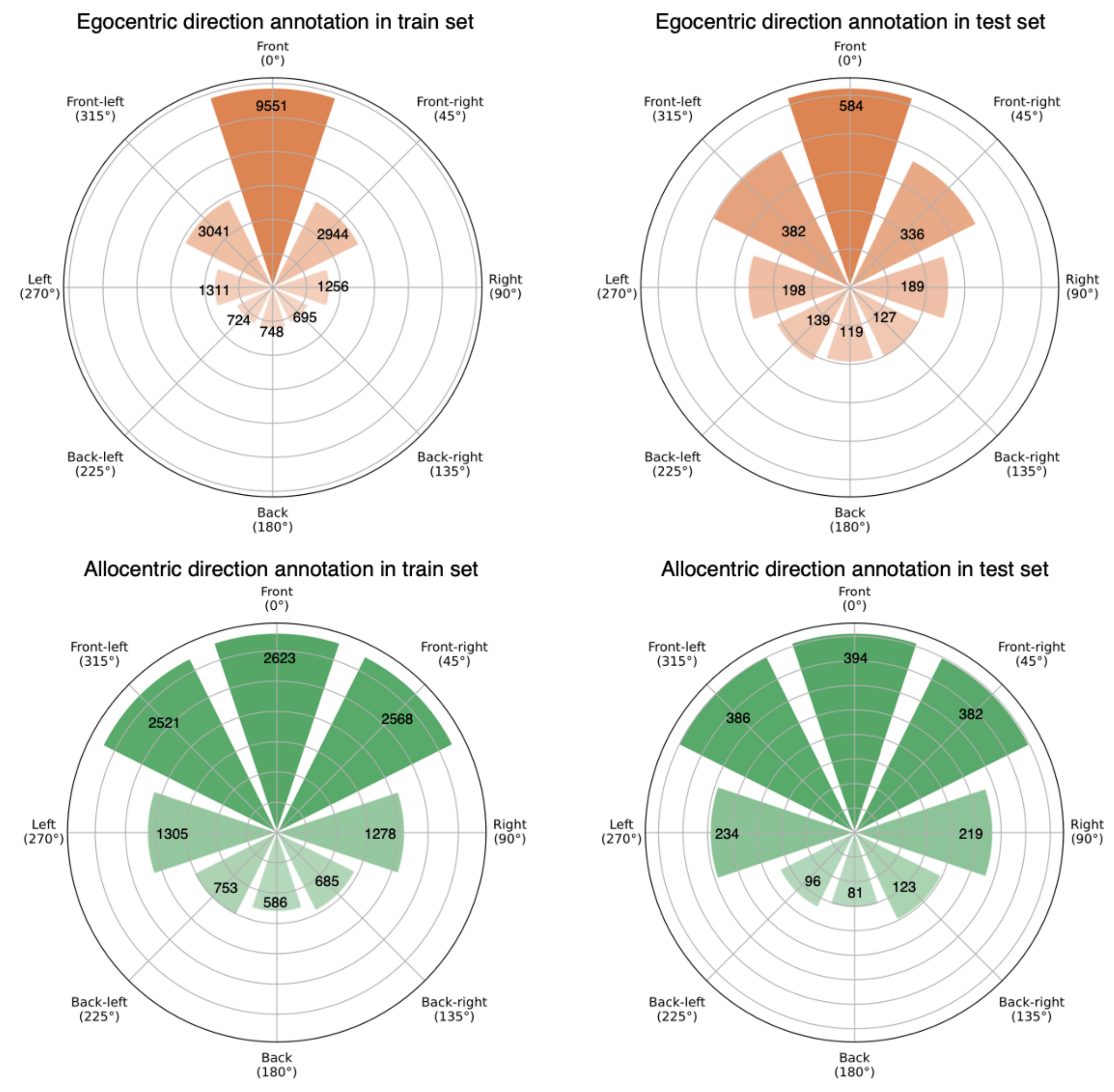}
    \caption{\textbf{Direction annotation statistics.} We show log-scaled distributions of direction-bin annotations for egocentric and allocentric perspectives in the training and test sets.}
    \label{fig:dir}
\end{figure}

\begin{figure}[t]
    \centering
    \includegraphics[width=1\linewidth]{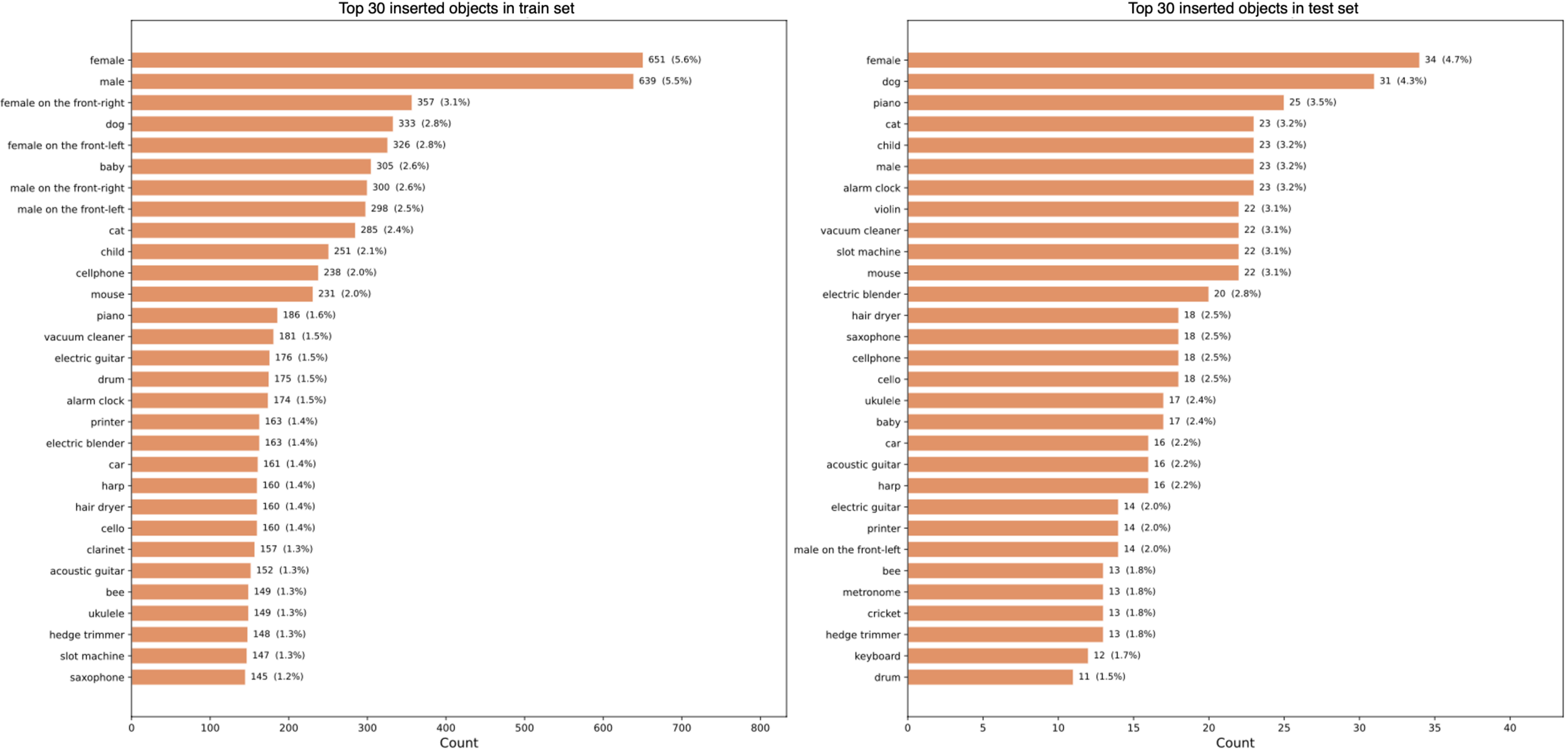}
    \caption{\textbf{Top 30 inserted object categories.} We show the 30 most frequent object categories inserted into training and test scenes.}
    \label{fig:insert}
\end{figure}

\begin{figure}[t]
    \centering
    \includegraphics[width=1\linewidth]{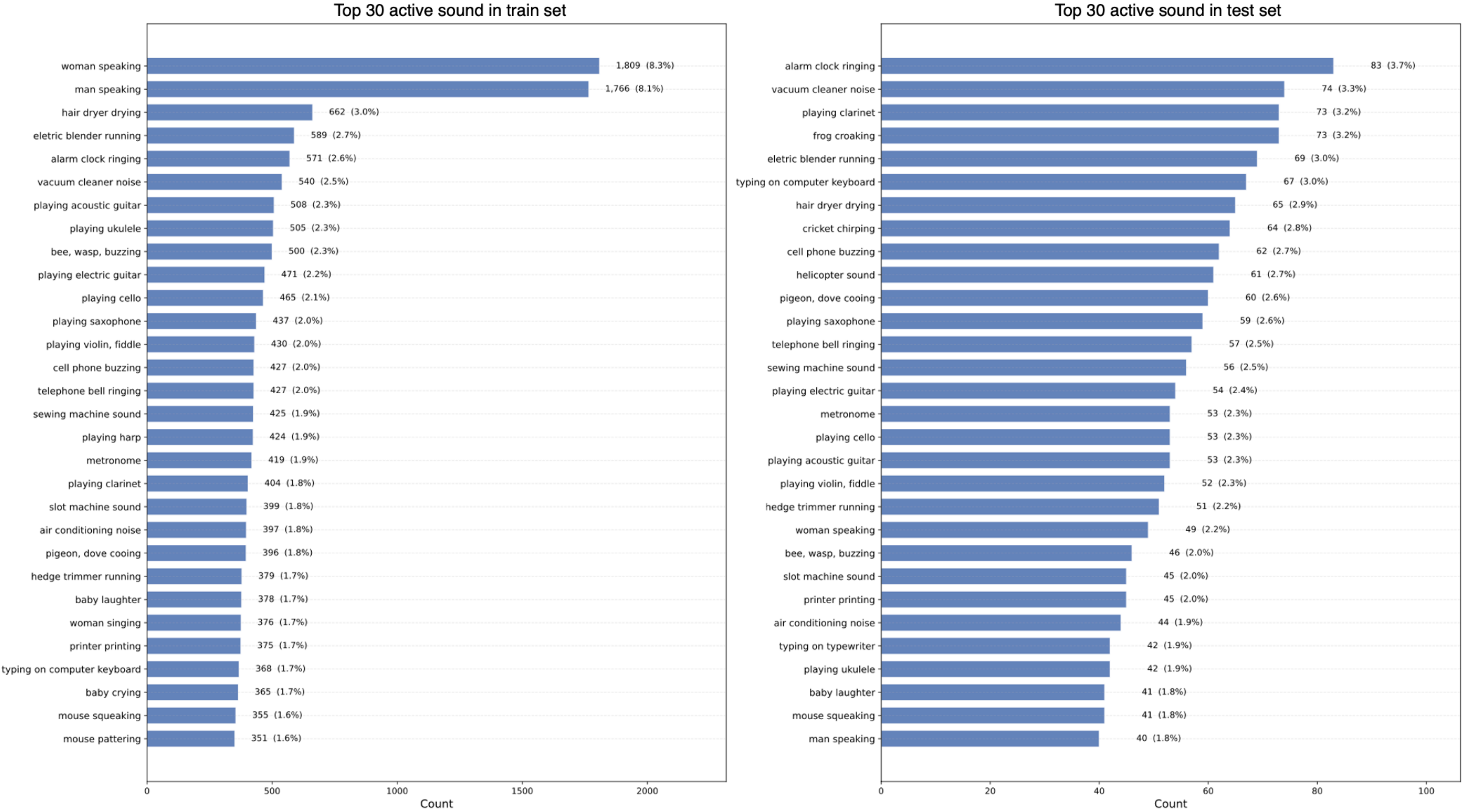}
    \caption{\textbf{Top 30 active sound categories.} We show the 30 most frequent sound categories used as active sound sources in training and test scenes.}
    \label{fig:audio}
\end{figure}

We use LLMs in three ways. First, the proposed HLR-LLM is built on a large language model as the reasoning module that integrates unified inputs, including visual, monaural audio, and binaural spatial audio tokens. Second, we use an LLM to paraphrase question templates to improve linguistic diversity and reduce the risk of models learning fixed template patterns. Finally, we use an LLM to assist with editing and grammar checking, table refinement (\Tref{tab:azimuth_distance_bins}), figure generation and refinement (\Fref{fig:dist},~\ref{fig:dir},~\ref{fig:insert},~\ref{fig:audio}), and development of the data-viewer web page.

\section{Licenses and terms of existing assets}\label{app:license}
We use existing datasets, simulators, models, and 3D assets only under their stated licenses and terms of use. 

\begin{itemize}
    \item \textbf{SoundSpaces 2.0}: used for spatial acoustic rendering. SoundSpaces is released under the Creative Commons Attribution 4.0 International license. Matterport3D-based task data and trained models follow the Matterport3D terms and CC BY-NC-SA 3.0 US license.
    \item \textbf{HM3DSem / HM3D}: used as the source of indoor 3D environments and semantic scene annotations. We use the dataset under its academic research terms and respect the restrictions of the underlying Matterport3D assets.
    \item \textbf{VGGSound}: used for non-speech sound events. We use the official split and retain only object-grounded categories suitable for indoor spatial reasoning. Copyright remains with the original video owners.
    \item \textbf{LibriSpeech}: used for speech audio and far-field ASR supervision. LibriSpeech is based on public-domain audiobooks and is distributed under a Creative Commons license.
    \item \textbf{Objaverse-XL}: used as a source of 3D object meshes. The dataset as a whole is licensed under ODC-By v1.0, while individual objects have their own licenses. We manually check and retain only assets compatible with our release setting.
    \item \textbf{Stable Diffusion 2}: used to generate canonical object images. The code is released under the MIT license, and model weights are released under the CreativeML Open RAIL++-M license.
    \item \textbf{TRELLIS}: used for image-to-3D mesh generation. TRELLIS models and code are released under the MIT license, with some submodules under their own licenses.
    \item \textbf{Comparison models}: MiniCPM-o-v2.6, VideoLLaMA2, Video-SALMONN~2, Qwen2.5-Omni, Qwen3-Omni, SAVVY, Hear You Are LLM, GPT-Realtime, and Gemini-2.5-Flash-Lite are used only for evaluation or baseline comparison under their respective model licenses, APIs, and terms of use.
\end{itemize}

All generated 3D assets are synthetic and are not intended to reproduce real people, private environments, or sensitive locations. We will release only assets and annotations that are compatible with the applicable licenses and redistribution terms.

\clearpage

\clearpage